\documentclass[10pt,twocolumn,letterpaper]{article}

\usepackage[pagenumbers]{cvpr} % To force page numbers, e.g. for an arXiv version

\usepackage[dvipsnames]{xcolor}
\usepackage{colortbl}
\usepackage{multirow} 
\usepackage{makecell}
\definecolor{mygreen}{RGB}{83,161,81}
\definecolor{myred}{RGB}{178,34,34}
\definecolor{lightorange}{RGB}{249,195,129}
\definecolor{borderblue}{RGB}{71,117,194}
\definecolor{borderyellow}{RGB}{253,190,38}

\newcommand{\Heng}[1]{\textcolor{black}{{}#1}}
\newcommand{\Mingfei}[1]{\textcolor{black}{{}#1}}
\newcommand{\MingfeiNew}[1]{\textcolor{black}{{}#1}}

\newcommand{\dataset}{PortraitMode-400\xspace}
\newcommand{\abbrdataset}{PM-400\xspace}

\newcommand{\app}{\raise.17ex\hbox{$\scriptstyle\sim$}}

\def\x{$\times$}
\newcolumntype{x}[1]{>{\centering\arraybackslash}p{#1pt}}

\newlength\savewidth\newcommand\shline{\noalign{\global\savewidth\arrayrulewidth
		\global\arrayrulewidth 1pt}\hline\noalign{\global\arrayrulewidth\savewidth}}

\makeatletter\renewcommand\paragraph{\@startsection{paragraph}{4}{\z@}
	{.5em \@plus1ex \@minus.2ex}{-.5em}{\normalfont\normalsize\bfseries}}\makeatother

\definecolor{cvprblue}{rgb}{0.21,0.49,0.74}
\usepackage[pagebackref,breaklinks,colorlinks,citecolor=cvprblue]{hyperref}

\def\paperID{**} % *** Enter the Paper ID here
\def\confName{CVPR}
\def\confYear{2024}

\title{Video Recognition in Portrait Mode}

\author{Mingfei Han$^{2,1,3}$\quad
Linjie Yang$^{1}$\quad
Xiaojie Jin$^{1}$\quad
Jiashi Feng$^{1}$\quad
Xiaojun Chang$^{2,4}$\quad
Heng Wang$^{1}$
\vspace{0.1em}\\
$^1$Bytedance \quad
$^2$ReLER Lab, AAII, UTS\quad
$^3$Data61, CSIRO \quad
$^4$MBZUAI\\
\url{https://mingfei.info/PMV/}
}

\begin{document}

\maketitle

%%%%%%%%% ABSTRACT
\begin{abstract}
%\Heng{The logic to follow when describing the story of the paper.}

%\Heng{Briefly describe that the current focus in video recognition has been on designing more powerful and/or efficient models/backbones. We argue that datasets have been playing a critical role to drive the progress in video recognition as well, from KTH to UCF101/Hollywood2 to Kinetics/Something-something. To some extent, the research problem we are studying is encapsulated in the datasets that we are working on. With the popularity of smart phone and social media applications, portrait mode videos have become an increasingly more popular format of video data (such as YouTube Shorts, Snapchat). There are new challenges that come with the new format of video data. Our paper looking into the challenges from the portrait mode videos, and provide thorough experimental results about the open research questions (such as the impact of aspect ratio to data augmentation, architecture design, inference; the importance of audio modality; bias of different video distribution) for portrait mode video recognition. Moreover, we designed a new dataset dedicated for portrait mode video recognition. We believe this dataset will inspire more researchers to work on portrait mode videos. }
%incubate

% The creation of new datasets often introduces new challenges to video recognition and inspires new ideas when tackling these challenges.
The creation of new datasets often presents new challenges for video recognition and can inspire novel ideas while addressing these challenges. 
%While existing datasets predominantly consist of landscape mode videos, our paper aims to introduce portrait mode videos to the research community and all the challenges that come with this format of video data. 
While existing datasets mainly comprise landscape mode videos, our paper seeks to introduce portrait mode videos to the research community and highlight the unique challenges associated with this video format.
%Portrait mode video recognition tasks are becoming increasingly more important due to the popularity of smart phones and social media applications.
With the growing popularity of smartphones and social media applications, recognizing portrait mode videos is becoming increasingly important.
%To this end, we build the first dataset, namely \dataset, dedicated for portrait mode video recognition.
To this end, we have developed the first dataset dedicated to portrait mode video recognition, namely \dataset.
%The taxonomy of \dataset is constructed in a data-driven way with XXX fine-grained categories and rigorous quality assurance is applied to ensure the correctness of human annotations. 
The taxonomy of \dataset was constructed in a data-driven manner, comprising 400 fine-grained categories, and rigorous quality assurance was implemented to ensure the accuracy of human annotations.
%Besides the new dataset, we thoroughly analyze the impact of video format (portrait mode \vs landscape mode) on recognition accuracy and the spatial bias due to different formats. 
In addition to the new dataset, we conducted a comprehensive analysis of the impact of video format (portrait mode versus landscape mode) on recognition accuracy and spatial bias due to the different formats.
%We also design extensive experiments as initial attempts to explore the key aspects of portrait mode video recognition, %designing vision models for portrait mode videos, 
%including the choice of data augmentation, %architecture design, 
%the importance of temporal information and audio modality. 
%We believe this dataset and our study will inspire the research community in this booming research field.
% Furthermore, we designed extensive experiments to explore key aspects of portrait mode video recognition, including the choice of data augmentation and evaluation procedure.
Furthermore, we designed extensive experiments to explore key aspects of portrait mode video recognition, including the choice of data augmentation, evaluation procedure, the importance of temporal information, and the role of audio modality.
 %With the insights from our experimental results and the introduction of our new \dataset dataset, our paper will inspire more research efforts in this new research direction.
 Building on the insights from our experimental results and the introduction of \dataset, our paper aims to inspire further research efforts in this emerging research area.

%We also carefully design experiments to investigate  such as data pre-processing, the importance of temporal information and to what extend audio is helpful 

%Most existing video recognition benchmarks focus on landscape mode videos mainly due to the abundant of this format of videos in the online video platforms. With the popularity of smart phones and social media, portrait mode videos have become an increasingly more popular format of video data. Such a video format poses new challenges for video understanding but lacks public benchmarks for algorithm development. We designed and collected a new portrait mode video recognition dataset named XXX with XX fine-grained human activity categories as a pioneering effort towards this direction of research. We also designed extensive experiments as the initial attempt to explore the key factors of designing vision models for portrait mode videos, including choice of data augmentation, architecture design, and importance of audio modality and temporal information. We believe this dataset and our study will inspire the research community in this booming research field.
\end{abstract}
\vspace{-2mm}

\section{Introduction}
%\Heng{I write a few bullet points for each paragraph. Please expand them into paragraphs. I will further polish the sentences later.}

%\Heng{Briefly summarize the progress we have made in the past two decades for video recognition. From the methodology perspective, we have evolved from local features + bags of features; to convolutional neural networks; to vision transformers. From the dataset perspective, we have evolved from controlled setting (\eg, KTH), to more realistic videos (\eg, Hollywood2, UCF101), to large-scale complex videos (\eg, Kinetics). Argue that dataset is a critical to drive research progress. We can quote ImageNet as an example.}

%Most efforts in video recognition have been focus on improving the accuracy and efficiency of different models and architectures on public benchmarks.
Most efforts in video recognition have focused on improving the accuracy and efficiency of different models and architectures on public benchmarks.
%During the past two decades, we have made dramatic shift in the types of video recognition models, % and tremendous progress of the recognition accuracy, 
%from bags of features~\cite{SivicZ03} to convolutional neural networks~\cite{XuYH15}, then more recent vision transformers~\cite{Arnab0H0LS21}. 
Over the past two decades, there has been a dramatic shift in the types of video recognition models, starting from bags of features~\cite{SivicZ03,wang2013action,peng2016bag,ullah2010improving,wang2013dense,peng2013hybrid,tang2013combining}, moving on to convolutional neural networks~\cite{XuYH15,KarpathyCVPR14,feichtenhofer2020x3d,wang2016temporal,feichtenhofer2019slowfast,wang2021tdn,yang2020temporal,lin2019tsm,tran2018closer,tran2019video,carreira2017quo}, and more recently, vision transformers~\cite{Arnab0H0LS21,arnab2021vivit,li2022uniformer,li2022mvitv2,fan2021mvit,bertasius2021space,liu2021videoswin,bulat2021space,liu2022swin,yuan2021tokens,patrick2021keeping}. 
%With the evolution of various models, video datasets have been a driving force behind the iteration of each generation of different models. 
With the evolution of various models, video datasets have played a crucial role in driving each generation of models.
The introduction of each video dataset has guided the research community to focus on new challenges. 
We have moved from using datasets collected in controlled environments (\eg, KTH~\cite{SchuldtLC04}, Weizmann~\cite{BlankGSIB05}) to more realistic videos (\eg, UCF101~\cite{UCF101}, HMDB51~\cite{KuehneJGPS11}), and now to large-scale web video datasets (\eg, Kinetics-700~\cite{kinetics700}, HowTo100M~\cite{miech19howto100m}).

\begin{figure}[t]
    % \small
    \centering
       \includegraphics[width=0.95\linewidth]{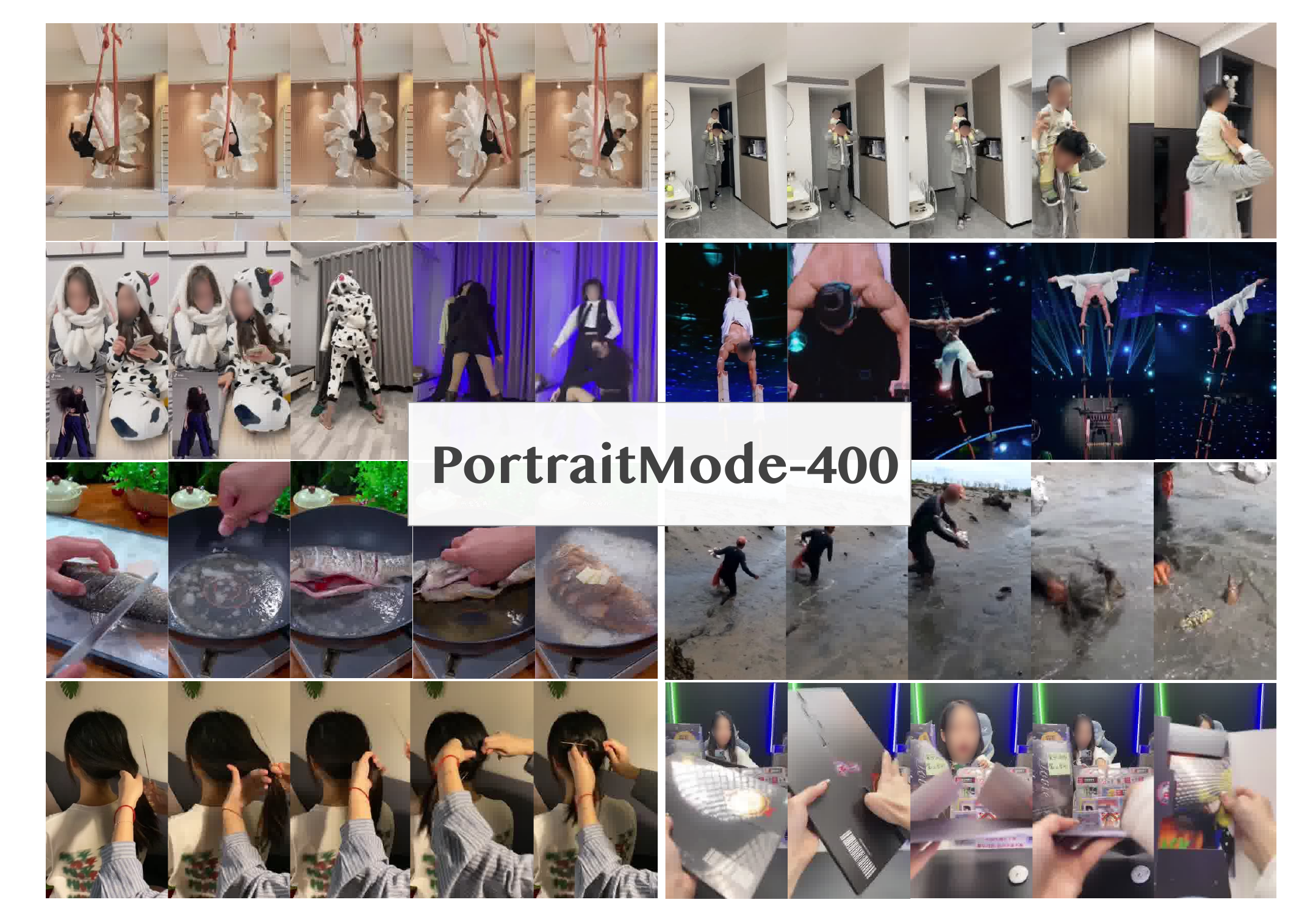}
    % % \vspace{-4mm}
    \caption{A glance of \dataset, which is the first dataset dedicated to portrait mode video recognition. It covers videos from 9 domains and 400 specific categories. We show video samples (left to right, top to down) for \textit{aerial yoga}, \textit{riding neck}, \textit{partner dancing (pop music)}, \textit{acrobatics}, \textit{cooking fish soup}, \textit{catching crab}, \textit{styling hair with hairpins} and \textit{opening mystery card packs}, from different domains of our dataset.
    }
    \label{fig:examples}
    \vspace{-3mm}
\end{figure}

%Existing video datasets are mostly built on landscape mode videos, which is the dominant video format on popular video websites, such as YouTube. 
%The popularity of smart phones and social media applications creates another trending video format, \ie, portrait mode videos. 
%On the mobile devices, portrait mode videos have taken the center stage on major social media applications.
While existing video datasets are mostly built on landscape mode videos, 
%it is worth noting that 
portrait mode videos have become increasingly more popular on major social media applications.
%, particularly on mobile devices.
%From landscape mode to portrait mode is not just a simple change of aspect ratio. It has deep implication about the types of content that are created, and different bias that comes with such data.
%However, 
The shift from landscape mode to portrait mode is not just %a matter of 
changing the aspect ratios of the videos. It has significant implications for the types of content that are created and the spatial bias inherent in the data.
Portrait mode videos bring in distinct challenges for video recognition as well. For example, they tend to focus more on the subject (\ie, typically humans) with much less background context, and include more egocentric content. In addition, they contain a lot of verbal communication that is essential to understand the video content.
%to understanding the content of the videos
%There is a great need of portrait mode video datasets in order to explore these new research problems.
%Therefore, 
There is a pressing need for portrait mode video datasets to explore these new research problems.

\begin{figure*}[t]
    \small
    \centering
       \includegraphics[width=0.9\linewidth]{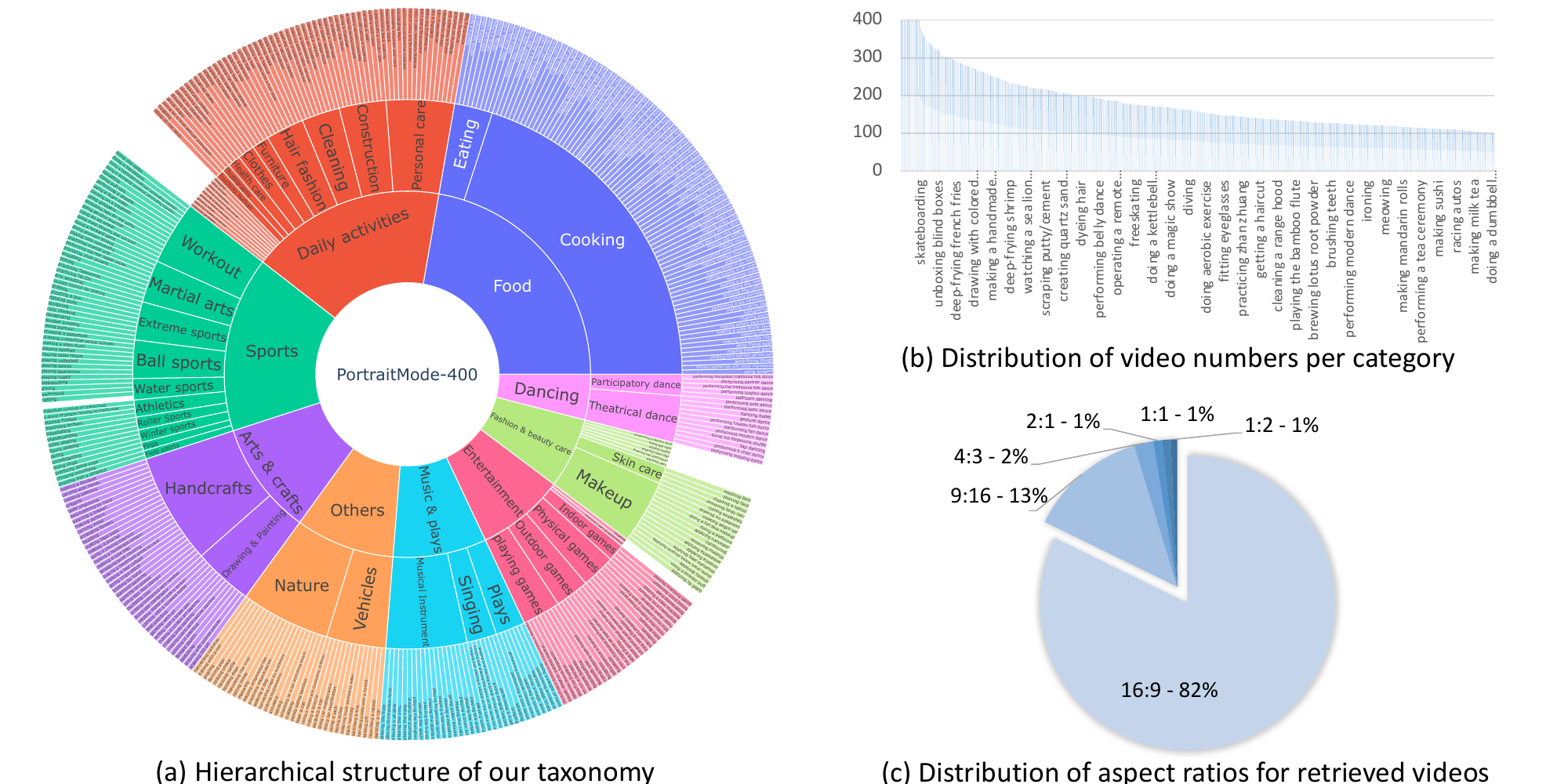}
    % \vspace{+2mm}
    \caption{Overview of our dataset. (a) We construct our taxonomy in a three-level hierarchical structure, which contains 9 domains and 400 leaf-node categories.
    % (b) We show the number of videos and categories in each of the super categories. The two subplots share the same horizontal axis for category names. It is shown that food, sports and daily activities together dominate our dataset, possessing over half of our videos and categories. Best seen in colour and with zoom.
    (b) We show the distribution of video numbers per category of our dataset, which contains a relatively balanced distribution of categories.
    (c) \Heng{We plot the distribution of aspect ratios for the retrieved videos via search queries.} The majority of videos (over 85\%) are in portrait mode, with 16:9 \Heng{being the dominant format}.
    }
    \label{fig:dataset}
    % \vspace{+2mm}
\end{figure*}

This paper introduces the first dataset dedicated to portrait mode video recognition, named \dataset (abbreviated as \abbrdataset), shown in Figure~\ref{fig:examples}. The dataset consists of 76k videos collected from Douyin\footnote{\label{fndy} Douyin is a popular social media application built for smartphones and primarily features portrait mode short-form videos. \url{https://www.douyin.com/}}, a popular short-video application, and annotated with 400 categories. The dataset's taxonomy is built in a data-driven way by aggregating search queries and covers a wide range of categories, including sports, food, music, handicrafts, and daily activities, among others. Many of the categories are fine-grained, as shown in Figure~\ref{fig:dataset} (a). The data annotation was performed by professionally trained human annotators, and additional quality assurance was conducted to improve the annotation accuracy and consistency. We built \dataset as a single-label dataset, and removed videos that can be tagged with multiple labels during annotation. While the recent 3Massiv~\cite{gupta20223massiv} dataset also includes a significant percentage of portrait mode videos, it is mostly built for multi-lingual and multi-modal research, and only has 34 coarse visual concepts, unlike \dataset.

% \wyt{maybe better to add one sentence includes the most important characteristics (potrait mode, multi-modal, short-term, action focused?) of the proposed dataset here or at the beginning of this paragraph or in the figure 1 caption.}

%, and organize our taxonomy in a three level tree structure (See Figure XX). Compared with a recent portrait video dataset 3Massiv~\cite{}, our dataset has larger scale ( $x\times$ ), more fine-grained action categories ( XX vs XX), and higher resolution (?).

%Besides introducing the \dataset dataset, we have also made initial attempts to explore several key research problems on portrait mode video recognition:

In addition to introducing the \dataset dataset, we have also made preliminary attempts to investigate several critical research problems related to portrait mode video recognition:

\begin{itemize}
\item How well does a model trained on landscape mode videos perform on portrait mode videos, and vice versa? 
%We construct a subset from Kinetics-700~\cite{kinetics700} for a rigorous comparison, and visualize the classification heatmaps (shown in Figure~\ref{fig:visualize_pm} and Figure~\ref{fig:visualize_lm}) to reveal the different spatial bias due to the change of video format.
We investigate this question by constructing a subset from the Kinetics-700 dataset~\cite{kinetics700} for a rigorous comparison and visualize classification heatmaps (shown in Figure~\ref{fig:visualize_pm} and Figure~\ref{fig:visualize_lm}) to reveal the differences in spatial bias resulting from the change in video format.
%We rigorously compare how does the mode generalize to 
\item %What are the best training and testing recipes for portrait mode video recognition?
What are the optimal training and testing protocols for portrait mode video recognition? 
%We dive into different components of modern deep learning systems, such as data augmentation, cropping strategies during evaluation, \etc.
We delve into various components of state-of-the-art deep learning systems, such as data augmentation, evaluation cropping strategies, \etc. 
%Some of our findings are contradicting to current standard practice for landscape mode videos, which indicates more research needs to be done for portrait mode videos.
Our discoveries challenge the existing conventions for landscape mode videos, thus necessitating further exploration into portrait mode videos.
%What kind of data augmentation is preferred for portrait mode videos when training action recognition models? Existing video-based CNNs and Transformers usually randomly resize the input video and then crop a square region for training, changing the aspect ratio of the video~\cite{}. Do we need to keep the aspect ratio during augmentation? 

%\item Do we need to modify the model architectures of CNNs and Transformers to accommodate the specific data format? Vision Transformers normally use square image sizes as input. Will adapting Transformers with a rectangular shape of input help improve the performance? 

\item\MingfeiNew{How important is temporal information for portrait mode videos? Can we recognize the actions from single frames~\cite{guo2014survey} or do we need to utilize temporal information for accurate results? 
% We investigate with various temporal utilization strategies. 
% Our findings reveal that strategically integrating temporal relations significantly enhances performance for video recognition in portrait mode.
We explore different temporal utilization strategies and find that integrating temporal information substantially improves video recognition in portrait mode.
% the importance of temporal information for portrait mode video recognition.
}
\item\MingfeiNew{Audio is a critical modality for video understanding~\cite{gao2020listen,kazakos2019epic}. Does audio contribute to video recognition in portrait mode? 
% Our experimental evidence indicates that even a basic integration of audio can yield substantial improvements in recognition accuracy, suggesting new avenues for multimodal video recognition.
Our experiments show that even simple audio integration can improve recognition accuracy, indicating possibilities for multimodal video analysis.
% This finding challenges existing perceptions and suggests new avenues for multimodal video analysis.
}

\end{itemize}

\section{Related Work}
%\noindent\textbf{Datasets for video recognition.} 
%Datasets often encapsulate the research problems that we want to investigate, especially for applications like video recognition. Some of the pioneering datasets for video recognition are collected in controlled setting, such as KTH~\cite{SchuldtLC04}, Weizmann~\cite{BlankGSIB05}, IXMAS~\cite{weinland2007action}, UIUC~\cite{tran2008human}, \etc. Their videos are often staged with simple and static background, and human actors are instructed to do scripted actions repeatedly. These datasets greatly simplify the action recognition problem and allow the model to focus on the action of interest. They inspire a series of works designing hand-crafted features~\cite{laptev2005space,laptev2008learning,klaser2008spatio,wang2013action} in combination with the bag of features model~\cite{SivicZ03,csurka2004visual}.
Datasets play a crucial role in investigating research problems, particularly in applications like video recognition. Several pioneering datasets for video recognition were collected in a controlled setting, including KTH~\cite{SchuldtLC04}, Weizmann~\cite{BlankGSIB05}, IXMAS~\cite{weinland2007action}, and UIUC~\cite{tran2008human}, \etc. The videos in these datasets are typically staged with simple and static backgrounds, and human actors are instructed to perform scripted actions repeatedly. By simplifying the action recognition problem, these datasets allow the models to focus on the action of interest. They have inspired the development of hand-crafted features~\cite{laptev2005space,laptev2008learning,klaser2008spatio,wang2013action} in combination with the bag of features models~\cite{SivicZ03,csurka2004visual}.

%Popular video websites (\eg, YouTube) quickly become the major source for creating video datasets. Unlike the aforementioned datasets, Internet videos are often more realistic and challenging with background clutter, camera motion, \etc.
Popular video websites, such as YouTube, have become the primary source of video datasets. Unlike the controlled datasets, Internet videos are more realistic and challenging due to factors like background clutter, camera motion, \etc.
%Many datasets are created by harvesting videos from websites like YouTube, \eg, UCF101~\cite{UCF101}, HMDB51~\cite{KuehneJGPS11}), Activitynet~\cite{heilbron2015activitynet}, Kinetics-400~\cite{kay2017kinetics}, Moments in Time~\cite{monfort2019moments}, \etc. They have become the major testbeds that support the development of many successful CNN architectures~\cite{simonyan2014two,tran2015learning,carreira2017quo,tran2018closer,feichtenhofer2019slowfast} and vision transformer~\cite{bertasius2021space,Arnab0H0LS21,liu2021videoswin} in the deep learning era. 
Several datasets are created by collecting videos from websites like YouTube, such as UCF101~\cite{UCF101}, HMDB51~\cite{KuehneJGPS11}, Activitynet~\cite{heilbron2015activitynet}, Kinetics-400~\cite{kay2017kinetics}, Moments in Time~\cite{monfort2019moments}, \etc. These datasets serve as the primary testbeds for the development of many successful CNN architectures~\cite{XuYH15,KarpathyCVPR14,feichtenhofer2020x3d,wang2016temporal,feichtenhofer2019slowfast,wang2021tdn,yang2020temporal,lin2019tsm,tran2018closer,tran2019video,carreira2017quo} and vision transformer models~\cite{Arnab0H0LS21,arnab2021vivit,li2022uniformer,li2022mvitv2,fan2021mvit,bertasius2021space,liu2021videoswin,bulat2021space,liu2022swin,yuan2021tokens,patrick2021keeping} in the deep learning era.
%There is a recent trend of building large-scale pre-training datasets (\eg, HowTo100M~\cite{miech19howto100m} and WebVid-10M~\cite{bain2021frozen}) with text supervision instead of labelled categories.
A recent trend is to build large-scale pre-training datasets, such as HowTo100M~\cite{miech19howto100m} and WebVid-10M~\cite{bain2021frozen}, using text 
% \wyt{replace text with natural language} 
supervision instead of labelled categories.

%Social media applications have been growing tremendously during the past few years. They have created in a new type of video data, \ie, portrait mode short-form videos, which are very different from the conventional landscape videos used in previous datasets. This inspires us to build a dataset dedicated for portrait mode videos and facilitate the research effort in this rising direction.
Social media applications have experienced tremendous growth in recent years, creating a new type of video data known as portrait mode short-form videos. These videos differ significantly from conventional landscape videos used in previous datasets, inspiring us to create %the creation of
a dataset dedicated to portrait mode videos.
%Note that the 3Massiv dataset~\cite{gupta20223massiv} also includes a large portion of portrait mode videos. However, it was intentionally designed for the multi-lingual and multi-modal purpose, and focuses on visual concepts instead of specific actions, with only 34 coarse concepts in total.
 It is worth noting that the 3Massiv dataset~\cite{gupta20223massiv} also includes a significant proportion of portrait mode videos. However, it was intentionally designed for multi-lingual and multi-modal purposes, focusing on visual concepts rather than specific actions, with only 34 coarse concepts in total.
 % \wyt{what is the difference between visual concepts and actions? static vs. dynamic?}

%We plot the the ratio of \#videos to the videos sampled using selected search queries. 

\section{The \dataset dataset}
% \Heng{We will briefly describe the process of constructing our dataset. Note that we may want to hide some sensitive details. We should keep this section concise.}
%In this section, we detail the process of constructing our \dataset dataset. First, we introduce our data-driven approach to build a taxonomy based on user queries. We then describe our annotation process and criteria we applied to ensure the annotation quality and consistency. Finally we compare \dataset with existing relevant datasets. 
In this section, we provide a comprehensive overview of the process behind constructing our \dataset dataset. We begin by discussing our data-driven approach to building a taxonomy, which is based on user queries. Next, we detail our rigorous annotation process and the criteria we applied to ensure high-quality and consistent annotations. Finally, we compare \dataset with existing datasets that are relevant to our work, highlighting the unique contributions and advantages of our dataset.

\subsection{Taxonomy}\label{sec:taxonomy}

% \Heng{We describe the process of collecting search queries (how do we decide whether a query is suitable or not, such as including verb, or other rules), then manually clean up the search queries, and recursively group the queries together in a bottom up way, and manually define the categories of the final taxonomy. We can provide slightly more details about how each step is done, what are the principals we try to follow when building the taxonomy.}

%\dataset is built using videos from a popular social media application Douyin\footref{fndy}, which is built for smartphones and mostly consists of portrait mode short-form videos. Instead of reusing categories from existing datasets, we build a new taxonomy for \dataset in order to better capture the different types of content that portrait mode videos can provide. We start with popular search queries from Douyin users, which often have text descriptions about the corresponding videos. However, search queries can be very noisy as well. In practice, we found many queries do not contain enough visual semantic meaning, such as celebrity names, song names, \etc. Our first step is to manually select candidate queries that contain verbs (\eg, \emph{eating cakes}) or nouns indicating potential actions (\eg, \emph{concealer} often leads to videos about how to using a concealer). We manually go through about 37,521 search queries and end up with 2420 usable queries that their corresponding videos may contain actions or motion, as we intend to make the final dataset includes more temporal information.

The videos in \dataset were sourced from Douyin\footref{fndy}. To better capture the various types of content that portrait mode videos can provide, we created a new taxonomy for \dataset instead of reusing categories from existing datasets. Our approach involved building the taxonomy based on popular search queries from Douyin users, which often include text descriptions about the corresponding videos. However, we found that many search queries lacked visual semantic meaning, such as celebrity names or song names. To address this, we manually selected candidate queries containing verbs (\eg, ``eating cakes'') or nouns indicating potential actions (\eg, ``concealer'' which often leads to videos about how to use a concealer). After manually examining approximately 38k search queries, we identified about 2.4k usable queries %that corresponded to 
with corresponding videos that might contain actions or motions, as we aimed to incorporate more temporal information in the final dataset.

\begin{table*}[t]
    \small
    \begin{center}
        
\setlength{\tabcolsep}{2.0mm}
\begin{tabular}
{@{} c   c   c   c   c   c   c}
\toprule

% Dataset & Ratio of PM & \#Classes & \#Videos & Duration & Avg. Duration & \#Videos/Class & Avg. \#videos & Year\\
% \arrayrulecolor{white}
% \arrayrulecolor{black}
% \arrayrulecolor{black}
% \arrayrulecolor{white}
% \arrayrulecolor{white}\hline
% \arrayrulecolor{black}\hline
% \arrayrulecolor{white}\hline
% % K700\cite{} & L+P & 700 & 66k & 1s-10s & 9s & 1-352 & 95 & Activity & '19\\
% S100-PM\cite{kinetics700} & 100\% & 100 & 20k & 1s-10s & 9s & 160-352 & 200 & '19\\
% \arrayrulecolor{white}\hline
% \arrayrulecolor{black}\hline
% \arrayrulecolor{white}\hline
% 3Massiv\cite{gupta20223massiv} & 95\% & 34 & 50k & 5s-2min & 20s & 578-4846 & 15k & '21\\
% \arrayrulecolor{white}\hline
% \arrayrulecolor{black}\hline
% \arrayrulecolor{white}\hline
% \dataset & 100\% & 400 & 56k & 2s-1min & 27s & 100-400 & 245 & '23 \\
% \arrayrulecolor{white}\hline
% \arrayrulecolor{black}\hline
% \arrayrulecolor{white}\hline

Dataset & \% of PM & \# of Classes & \# of Videos & Duration & Avg. Duration & Year\\
% \arrayrulecolor{white}
% \arrayrulecolor{black}
% \arrayrulecolor{black}
% \arrayrulecolor{white}
% \arrayrulecolor{white}\hline
% \arrayrulecolor{black}\hline
% \arrayrulecolor{white}\hline
\midrule
% K700\cite{} & L+P & 700 & 66k & 1s-10s & 9s & 1-352 & 95 & Activity & '19\\
S100-PM \cite{kinetics700} & 100\% & 100 & 20k & 1s-10s & 9s & '19\\
% \arrayrulecolor{white}\hline
% \arrayrulecolor{black}\hline
% \arrayrulecolor{white}\hline
3Massiv \cite{gupta20223massiv} & 95\% & 34 & 50k & 5s-2min & 20s & '21\\
% \arrayrulecolor{white}\hline
% \arrayrulecolor{black}\hline
% \arrayrulecolor{white}\hline
\midrule
\dataset & 100\% & 400 & 76k & 2s-1min & 27s & '23 \\
% \arrayrulecolor{white}\hline
% \arrayrulecolor{black}\hline
% \arrayrulecolor{white}\hline
\bottomrule

% 720 x 1280 
% 1. random ratio -> 10\% pixel -> pixel number
% 2. crop shape -> rectangle
% 3. resize -> 224x224

\end{tabular}
    \end{center}
    \vspace{-4mm}
    \caption{Comparison of different portrait mode video datasets. S100-PM is a portrait-mode-only subset sampled from Kinetics-700, as detailed in Section~\ref{sec:comparison_datasets}. 3Massiv contains 5\% landscape mode videos and is targeted for video classification in 34 coarse categories. Our \dataset contains portrait mode videos only and has more videos in a diversified taxonomy (400 classes).}
    \label{tab:dataset}
    % % \vspace{-1mm}
\end{table*}

%With the initial selected search queries, our second step is to aggregate the queries recursively in a bottom-up way. This process can produce increasingly more abstract concepts and result in a hierarchical tree structure taxonomy as shown in Figure~\ref{fig:dataset} (a). Besides the final taxonomy, there are two additional goals we are trying to achieve in the second step: 1), merge similar queries to a final leaf node category of the taxonomy; 2) split or remove the queries that may overlap with existing categories so that all the final categories are mutually exclusive. 
With the initial set of selected search queries, our second step is to recursively aggregate the queries in a bottom-up manner. This process generates increasingly abstract concepts, resulting in a hierarchical tree structure taxonomy, as illustrated in Figure~\ref{fig:dataset} (a). In addition to producing the final taxonomy, we have two other objectives in this step: 1) merging similar queries into a final leaf node category of the taxonomy; 2) splitting or removing queries that may overlap with existing categories, so that all final categories are mutually exclusive.
% \footnote{\Heng{Mingfei, please come up with an example of splitting or removing queries}}
For example, we merge \emph{tutorials for fitness}, \emph{exercises for weight loss} and \emph{fat-burning fitness exercises} to \emph{aerobics}; 
\Mingfei{we split \emph{calligraphy exercise} into \emph{pen calligraphy} and \emph{brush calligraphy}}.
%After the second step, 501 candidate categories are generated from 2420 selected search queries, and they form a three-layer hierarchy as illustrated in Figure~\ref{fig:dataset} (a).
After completing the second step, we obtained about 500 candidate categories derived from the 2.4k selected search queries, which are organized in a three-layer hierarchy as depicted in Figure~\ref{fig:dataset}.% (a).

The taxonomy used in the Kinetics-400~\cite{kay2017kinetics} dataset is built through a combination of reusing categories from previous datasets and crowdsourcing. %However, 
In contrast to Kinetics-400, our taxonomy is developed using a data-driven approach that better reflects the current trends in social media. Besides, our taxonomy covers a wider range of content, including everyday activities (\emph{food}, \emph{beauty care}, \emph{entertainment}, \etc)
% \wyt{(food or cooking? is food an activity? or daily scenes?)}
, natural phenomena (\emph{raining}, \emph{snowing}, \etc) as well as transportation-related activities (\emph{airplane taking off}, \emph{launching rocket}, \etc). This is in contrast to existing datasets that mostly focus on human actions.
% \wyt{only}. 
Furthermore, our taxonomy offers more fine-grained categories compared to 3Massiv~\cite{gupta20223massiv}, which is designed for coarse visual concept classification. For instance, while 3Massiv has only one class for food, our taxonomy includes 89 distinct categories under the food parent node, covering various types of food and food-related activities such as cooking and eating.

%As shown in Fig. xx, our taxonomy is constructed hierarchically with a three-level tree structure. Different from Kinetics, our taxonomy is designed to include not only human activities but also other contents that people care about, such as natural events and transportation. Sourced from popular search queries, our taxonomy focuses more on daily life, such as food, beauty care, sports and entertainment. Compared to 3Massiv, whose taxonomy is designed for coarse action classification, our taxonomy is dedicated to fine-grained video content understanding. For example, 3Massiv has a single class for food while ours has 50 categories under the food super-category, including cooking and eating various foods. 

\subsection{Sampling and annotation}

% \Heng{For each given category or query, how do we sample videos for that category? We randomly sample videos that have been viewed by xxx times, we should limit the length of the video to xx seconds to xx seconds. How do we make sure that the videos are portrait mode videos?
% We should also briefly describe the deduplication process.}

For each of the \Mingfei{500} candidate categories in the taxonomy, we have about \Mingfei{2 to 50} selected search queries associated with it, as described in Section~\ref{sec:taxonomy}. We retrieve \Mingfei{1.2k to 740k} videos for each query from Douyin\footref{fndy} depending on how frequently the query has been searched. Subsequently, we create a pool of videos for each category by aggregating all the retrieved videos from their corresponding queries. Figure~\ref{fig:dataset} (c) illustrates the distribution of the aspect ratio of the retrieved videos. Although 16:9 is the dominant aspect ratio, there are also other aspect ratios for portrait mode videos, such as 4:3. For the video pool, we use a few criteria to sample target videos for annotation: 1) we select videos whose aspect ratios (\Mingfei{height/width}) are greater than 1 to ensure that \dataset includes only portrait mode videos; 2) we select videos whose duration is shorter than 1 minute to limit annotation costs; and 3) we select videos that have been viewed over 700 times by Douyin users to ensure that our dataset better reflects the typical types of content for portrait mode videos.

%Finally, we conduct deduplication to remove duplicated or similar videos in the video pool. Specifically, we extract feature vectors of each video using a Uniformer-base model~\cite{li2022uniformer} pre-trained on Kinetics-700 dataset~\cite{kinetics700}. We build a graph by connecting video pairs whose feature vectors have a cosine similarity greater than 0.98. We perform the Louvain algorithm~\cite{blondel2008fast} on the graph to find video clusters, and then discard all the videos except one in each cluster.
%About 25.2\% of videos are been removed through deduplication. Videos met all the above criteria then enter the next stage for human annotation.

Finally, we perform deduplication on the video pool to eliminate duplicated or similar videos. To achieve this, we extract feature vectors of each video using Uniformer-Base~\cite{li2022uniformer} pretrained on Kinetics-700 dataset~\cite{kinetics700}. Next, we build a graph by connecting video pairs with feature vectors having a cosine similarity greater than 0.98. We then apply the Louvain algorithm~\cite{blondel2008fast} on the graph to identify video clusters and discard all the videos in each cluster except one. About 25\% of videos are removed through deduplication, and only videos that meet all the aforementioned criteria move on to the next stage for human annotation.

%With the above taxonomy, we have one or multiple candidate search queries for each class. We search videos from Douyin with the candidate queries and further merge them to obtain candidate videos for each class. We remove videos whose aspect ratios (height/width) are greater than 1 to ensure they are in portrait mode. We also filter the length of videos by keeping videos shorter than 1 minute. Finally, we conduct deduplication to remove duplicate or similar videos. Specifically, we extract feature vectors of the videos by a Uniformer-base model pretrained on Kinetics-700 dataset and construct a graph by connecting video pairs with cosine similarity greater than 0.98. We perform the Louvain algorithm~\cite{} on the graph to find video clusters, and then discard all the videos except one in a cluster.

% \Heng{Briefly describe the annotation process: the annotator is asked to do a binary decision to check the label is a good match to the video content. We also cross check whether there are multiple labels can be applied on a single video. We reject those examples to insure our dataset is a single label dataset. We can offer some basic statistics about the percentage of videos that are rejected.  How do we insure the annotation quality? Each annotator are first trained before starting annotation. We also do quality assurance to check the correctness of the annotated videos.}

The human annotation task is straightforward. An annotator is presented with a given category and its video pool, and is asked to confirm or deny whether the category name is a good match for the content of each video. Before starting annotation, annotators undergo training to learn the annotation criteria for all candidate categories, and they are required to pass a quality check test. Only annotators with an accuracy greater than 95\% are qualified for annotation to ensure the accuracy and consistency of their annotations. During annotation, annotators %are asked to 
discard videos that may be confused with multiple categories of our taxonomy, ensuring that \dataset is a strictly single-label dataset. Under our restricted rules, approximately 65\% of videos are rejected. To ensure annotation quality, approximately 20\% of annotations are reviewed by two additional examiners.

%The resulted videos are then annotated by a group of annotators. Each annotator is given a list of video-class pairs and is asked to answer a binary question for each pair, i.e., whether the video and the class label are a good match. Before annotation, annotators are trained on the annotation criteria of all candidate classes and are required to pass a quality check test. Only annotators achieving accuracy greater than 95\% are qualified for annotation. During annotation, the annotators are asked to discard videos that match with multiple classes covered by our taxonomy. Under the restricted rules, around 65\% of the candidate videos are rejected. To ensure high annotating quality, 20\% of the annotated video-class pairs are reviewed and assessed by two trained quality examiners.

% \subsection{Compare \dataset with existing datasets}
\subsection{Comparisons with existing datasets}
\label{sec:comparison_datasets}

% \begin{figure}[t]
%     \small
%     %\fbox{\rule{0pt}{2in} \rule{0.9\linewidth}{0pt}}
%        \includegraphics[width=0.98\linewidth]{figures/ours_taxonomy.pdf}
%     % % \vspace{-4mm}
%     \caption{The hierarchical taxonomy of our dataset. It is designed in a three-level tree structure. The 9 top categories can cover all the video content in our dataset.}
%     \label{tab:dataset}
%     % % \vspace{-2mm}
% \end{figure}

% \begin{figure*}[t]
% \begin{center}
% %\fbox{\rule{0pt}{2in} \rule{0.9\linewidth}{0pt}}
%    \includegraphics[width=0.98\linewidth]{figures/video_number_distribution.pdf}
% \end{center}
% % \vspace{-6mm}
% \caption{Video distribution. \Mingfei{Not in a long-tail distribution. Overall view of our dataset distribution.}}
% \label{fig:visualize_vid_dist}
% \end{figure*}

% \Heng{Present the final statistics of the dataset, such as number of videos, number of categories, number of videos per categories. Mention that we limit the number of videos for each category to be in a range xxx to xxx, so that the dataset is not a long tail distribution. Compare our datasets with existing other datasets (such as the CVPR 2022 dataset, and portrait mode Kinetics-700 subset).}

After finishing annotating all the videos, we keep all the categories that have at least 100 videos. We keep at most 400 videos per category so that the distribution of videos across different categories are more or less balanced, as shown in Figure~\ref{fig:dataset} (b). Our dataset contains 76k videos in total, spanning over 400 categories. 
% We do plan to continue the annotation and make our dataset even larger. 
We randomly sample 50 videos per category for testing, and the rest are used for training. Table~\ref{tab:dataset} compares the statistics of \dataset with other relevant datasets.
Though 3Massiv mostly includes portrait mode videos, it is a multi-lingual and multi-modal dataset designed for concept recognition with only 34 coarse concepts. 
\dataset has a more diversified and fine-grained taxonomy that is dedicated for portrait mode video recognition.

%As shown in Figure~ref{fig:dataset}, our dataset contains 56k videos spanning over 400 categories. We limit the number of videos for each category to be in a range of 100 to 400, to prevent a long-tail distribution of different classes. 
% In the current literature, there are two related datasets, Kinetics-700 and 3Massiv. 
%Compared to the current related datasets, our dataset achieves a trade-off between class balance and taxonomy diversity.
%Kinetics-700 has the most annotated portrait mode videos (68k), whereas the number of videos per category is in a long-tail distribution, ranging from 1 to 362. 
%3Massiv is dominated by portrait mode videos, which is designed for social media video concept recognition in 34 coarse categories. 
%Differently, our dataset has a more diversified and fine-grained taxonomy that is dedicated to portrait mode video recognition. 

% \Heng{We also build a portrait mode version from the Kinetics-700 datasets. Please describe the process of how do we build this Kinetics-700 portrait dataset. And how we can use it to compare portrait mode dataset vs. landscape mode dataset. Mention that we also try to use software (cite google's tool here) to convert landscape mode video to portrait mode, but found the data quality is not satisfactory. In the end, building portrait mode Kinetics-700 is the best choice we have right now to do rigorous comparison between landscape mode and portrait mode.}

To conduct a rigorous comparison between landscape mode and portrait mode video recognition, we created two subsets from the Kinetics-700 dataset: a portrait mode subset and a corresponding landscape mode subset. The details of these subsets are shown in Table~\ref{tab:dataset}.
%We build the portrait mode subset using the top 100 categories with most portrait mode videos in Kinetics-700, named as Selected-100 Portrait Mode (S100-PM).
%Each category in S100-PM has 160 to 352 portrait mode videos, resulting in a total of 20k videos.
We first constructed the portrait mode subset, named {\bf Selected-100} Portrait Mode (S100-PM), using the top 100 categories with the most portrait mode videos in Kinetics-700. Each category in S100-PM contains 160 to 352 portrait mode videos, resulting in a total of 20k videos.
%Using the same categories from S100-PM, we further build a counterpart landscape mode version from Kinetics-700, named Selected-100 Landscape Mode (S100-LM), by sampling the same number of landscape mode videos as the S100-PM for each category.
To build a counterpart landscape mode version from Kinetics-700, we sampled the same number of landscape mode videos as S100-PM for each category, resulting in a landscape mode subset named Selected-100 Landscape Mode (S100-LM).
%Thus S100-PM and S100-LM have an identical taxonomy and the same video distribution per category.
Therefore, S100-PM and S100-LM have the same taxonomy and the same video distribution per category.
%Though the video content of S100-PM and S100-LM can be different due to different video formats,
Although the video content of S100-PM and S100-LM may differ due to different video formats,
%are different may be  different have a slight domain difference due to the natural distribution of portrait mode and landscape model videos, 
%we argue that they are still useful benchmarks %can still be promising tools 
%to illustrate and validate the difference between landscape mode and portrait mode video recognition.
we believe that they are still useful benchmarks for illustrating and validating the difference between landscape mode and portrait mode video recognition.
We have also tried AutoFlip\footnote{\url{https://ai.googleblog.com/2020/02/autoflip-open-source-framework-for.html}} %(a saliency-based cropping software from Google) 
to convert landscape mode videos to portrait mode, thereby ensuring the same video content in both subsets.
%Unfortunately, converted portrait mode videos have %, which resulted in 
%unsatisfactory data quality.
However, the converted portrait mode videos had unsatisfactory data quality.
%To this end,
%Till now, building S100-PM and S100-LM from Kinetics-700 is the best choice %we have for 
%to rigorously compare different video formats on recognition tasks.
%between the different display modes.
Thus, building S100-PM and S100-LM from Kinetics-700 remains the best option for rigorously comparing different video formats on recognition tasks.

% \input{sections/method}
%\section{Experiments}

%\section{Experimental setup}
%\Heng{Not sure we need this section. I think we can integrate this into section 2.3.}

%We define the videos with aspect ratios (height/width) greater than 1.75 as Portrait Mode (PM) videos and the videos with aspect ratios smaller than 0.76 as Landscape Mode (LM) videos. We use the following datasets as the primary benchmarks.

%\textbf{3Massiv} \cite{gupta20223massiv} consists of 50k annotated short videos in 34 different content categories, in which 32k videos are used for training, 10k videos for validation and others for testing. 92\% videos of 3Massiv are in portrait mode.

%\textbf{Selected-100} is sourced from Kinetics-700 with selected 100 categories. It has two versions containing videos in different display modes, \ie S100-PM with videos in PM and S100-LM with videos in LM. For both versions, the dataset has 15k videos for training and 5k videos for validation.

% \textbf{DYO} contains xxk videos in 80 categories, which are in portrait mode and centrally hosted by Bytedance.

\begin{table}[t]
    \small
    \begin{center}
        
\setlength{\tabcolsep}{2mm}
% \tablestyle{2pt}{1.05}
\small
\begin{tabular}
{@{} l | c | c | c | c}

% shorter - single crop test
% \toprule
% \multirow{2}{*}{Method} & \multirow{2}{*}{Train set} & \multicolumn{2}{c}{Top-1 Acc.} & \multirow{2}{*}{GFLOPs\x views}\\
\multicolumn{1}{l|}{Model} & Train  & Val. &   Acc.  & GFLOPs\x views  \\
\shline
\multirow{4}{*}{X3D-M\cite{feichtenhofer2020x3d}}& \multirow{2}{*}{PM}  & \cellcolor{gray!20} PM & \cellcolor{gray!20} \textbf{52.0}  & \cellcolor{gray!20} 4.9$\times$3$\times$10 \\
 &   & LM & 41.2 & 4.9$\times$3$\times$10 \\
% \arrayrulecolor{white}\cline{2-6}
\arrayrulecolor{black}\cline{2-5}
% \arrayrulecolor{white}\cline{2-6}
 & \multirow{2}{*}{LM}  & PM & \textbf{44.5} & 4.9$\times$3$\times$10 \\
 &   & LM & 43.5 & 4.9$\times$3$\times$10 \\
\hline
% \rowcolor{gray!20} 
\multirow{4}{*}{Uniformer-S\cite{li2022uniformer}} & \multirow{2}{*}{PM} & \cellcolor{gray!20} PM  & \cellcolor{gray!20} \textbf{42.0} & \cellcolor{gray!20}  41.8$\times$1$\times$4 \\
 &   & LM  & 36.2 & 41.8$\times$1$\times$4 \\
\arrayrulecolor{black}\cline{2-5}
 & \multirow{2}{*}{LM} & PM  & 40.1 & 41.8$\times$1$\times$4 \\
 &   & LM  & \textbf{40.8} & 41.8$\times$1$\times$4 \\
\hline

\multirow{4}{*}{MViTv2-S\cite{li2022mvitv2}} & \multirow{2}{*}{PM} & \cellcolor{gray!20} PM & \cellcolor{gray!20} \textbf{41.0} & \cellcolor{gray!20}  64.0$\times$1$\times$5 \\
% \multirow{2}{*}{MViTv2-S}    & PM & 36.94 & 32.16 \\
% \arrayrulecolor{white}\cline{2-4}
% \arrayrulecolor{black}\cline{2-4}
% \arrayrulecolor{white}\cline{2-4}
                       &   & LM & 35.7 & 64.0$\times$1$\times$5\\
\arrayrulecolor{black}\cline{2-5}
                        & \multirow{2}{*}{LM} & PM & 33.7 & 64.0$\times$1$\times$5\\
                        &   & LM & \textbf{36.3} & 64.0$\times$1$\times$5\\

\end{tabular}
    \end{center}
    \vspace{-6mm}
    \caption{Cross mode evaluation with different models on Selected-100. Evaluation results performed on the PM subset correspond to the last column of Table~\ref{tab:ab_res}. Views during inference are shown by the multiplication of \# of spatial crops and \# of temporal views. Rows highlighted perform best for the corresponding model.}
    % , \ie, models trained with LM and PM videos are evaluated with videos in same display mode.}
    \label{tab:ab_s100}
    % % \vspace{-2mm}
\end{table}

\section{Landscape Mode \vs Portrait Mode}
Landscape and portrait mode videos, often shot in different ways and purposes, display unique content and biases. This affects subjects' action patterns and overall visual dynamics. Therefore, models trained on one mode may struggle in the other. This section examines how models adapt across these different modes, focusing on their spatial information and cross-mode generalizability.

\subsection{Cross Mode Evaluation}
% \Mingfei{We need to describe the setup of cross-mode learning first. Why did we design such an experiment scheme? Why is such a design reasonable?}

%Shot in different ways, the movement range of the actor in the video is constrained in different directions, either vertically or horizontally.
%The aspect ratio of a video will interfere with the content in a video. 
To show the impact of the different domain priors of landscape and portrait mode videos on video recognition tasks, comparisons need to be made between the same video content shot in portrait mode and landscape mode.
%For the example that a man is climbing a tree, landscape mode videos usually contain much background on the two sides of the video and adjust the view to follow the actor, while portrait mode videos can easily capture the movement and could keep the view unchanged.
%Ideally, the experiments should be performed on the videos shot two times for the same actor moving, which is infeasible to achieve.
Ideally, for each action or event, we should shoot it with both portrait mode and landscape mode cameras. However, such a process is time-consuming and hard to achieve.
%We turn to a compromise to sample landscape mode videos and portrait mode videos in the same taxonomy.
%One alternative solution we consider is to transform the current landscape mode video dataset (e.g., Kinetics \cite{kay2017kinetics}) into portrait mode. However, this approach results in unsatisfactory video quality. 
Therefore, we opt for sampling original portrait mode videos and landscape mode videos with the same distribution and taxonomy from Kinetics-700 \cite{kinetics700}, as detailed in Section~\ref{sec:comparison_datasets}.

% \begin{figure}[t]
% \begin{center}
% %\fbox{\rule{0pt}{2in} \rule{0.9\linewidth}{0pt}}
%    \includegraphics[width=1.0\linewidth]{figures/LM_PM_results_5cats_v6.pdf}
% \end{center}
% % % \vspace{-5mm}
% \caption{Top five categories with the largest accuracy improvement and largest accuracy degradation with Uniformer-S for cross mode comparison. %trained on PM and LM videos. 
% %(a) Improvement of training on PM compared to training on LM. (b) Improvement of training on LM compared to training on PM.
% }
% \label{fig:visualize_lm_pm_5cats}
% \end{figure}

%To demonstrate the differences in data priors and biases between portrait mode videos and landscape mode videos, \Mingfei{we evaluate models trained with different training sets on the same validation set.}
To explore the impact of the different priors to video recognition models, we conducted extensive experiments using different subsets of S100 (S100-PM and S100-LM).
%, which have the same semantic taxonomy and number of videos per class. 
% Our experiments involve training models on different subsets and then evaluating their performance on either portrait or landscape mode videos. 
% We randomly select 25\% videos as test set for each subset respectively.
We trained various models on different subsets and evaluated their performance on landscape mode videos and portrait mode videos, by randomly selecting 25\% videos as the validation set for each subset.
For example, evaluated on S100-PM, models trained with S100-PM and S100-LM respectively can be fairly compared to see which video type is more effective to train models for videos in portrait mode.
We conduct the experiments on three models, i.e. a CNN model X3D \cite{feichtenhofer2020x3d}, a hybrid transformer model Uniformer \cite{li2022uniformer}, and a pure transformer model MViTv2 \cite{li2022mvitv2} to show the impact of video formats %training data in the two domains 
on different model architectures. 
During training and testing, we resize frames based on the shorter side while preserving aspect ratios and crop them into 224$\times$224 pixel squares for input.
We train all models from scratch without pretraining to avoid the impact of pretraining dataset. Popular pretraining datasets such as ImageNet~\cite{krizhevsky2012imagenet} are biased towards landscape images which may add additional bias to our analysis.

We summarize all results as in Table~\ref{tab:ab_s100}.
%we experiment with different models on different training and evaluation settings on S100.
%We can see that X3D trained with \Mingfei{PM videos shows a large performance gain evaluated in the two datasets compared with X3D trained with LM videos}.
%The same phenomenon can be observed for Uniformer and MViT respectively.
By comparing results \Heng{in each row,} % horizontally, 
we find that models trained on PM videos has a larger performance gap on the PM and LM testsets than models trained with LM videos. 
% By validation on the identical subset, it can be inferred that videos in different display modes contain different data bias.
Moreover, %when comparing results vertically in the same column, we can see that 
models trained on PM data usually have better performance on PM testset compared to the models trained with LM videos. For example, \Mingfei{evaluated on S100-PM, X3D trained with PM videos outperforms the model trained with LM videos by a large margin} of $~8\%$ (51.2\% vs. 44.5\%).
When evaluated on S100-LM, X3D achieves relatively comparable performance either trained with PM videos or LM videos (41.2\% vs. 43.5\%).
This indicates that \Mingfei{training videos in portrait mode are necessary to achieve satisfying performance on portrait mode videos}.

\subsection{Spatial priors}
\label{sec:pm_lm_vis}
To investigate the different spatial data priors of portrait mode videos and landscape mode videos, we extensively evaluate the models trained on S100-PM and S100-LM on different frame positions to show the importance of frame features at different locations.

Specifically, \Mingfei{we first train Uniformer-S \cite{li2022uniformer} with 112$\times$112 crops and shorter-side resized (set to a random value between 256 and 320) frames on either S100-PM or S100-LM. We name the resulted two models Probing-P and Probing-L.
Then we evaluate the models with crops of 112$\times$112 on different locations in a sliding window at the shorter-side resized video clips.}% at a 224x? or ?x224 image resolution}. %take the S100-PM validation set as an instance, 
\Mingfei{
The sliding strides vary for portrait mode and landscape mode videos in both height and width.
For portrait mode videos, the stride in height is set to 1/16 of the frame height and the stride in width is set to 1/9 of the frame width. 
Sliding strides of landscape mode videos are adjusted vice versa.}

\begin{figure}[t]
\begin{center}
   \includegraphics[width=1.0\linewidth]{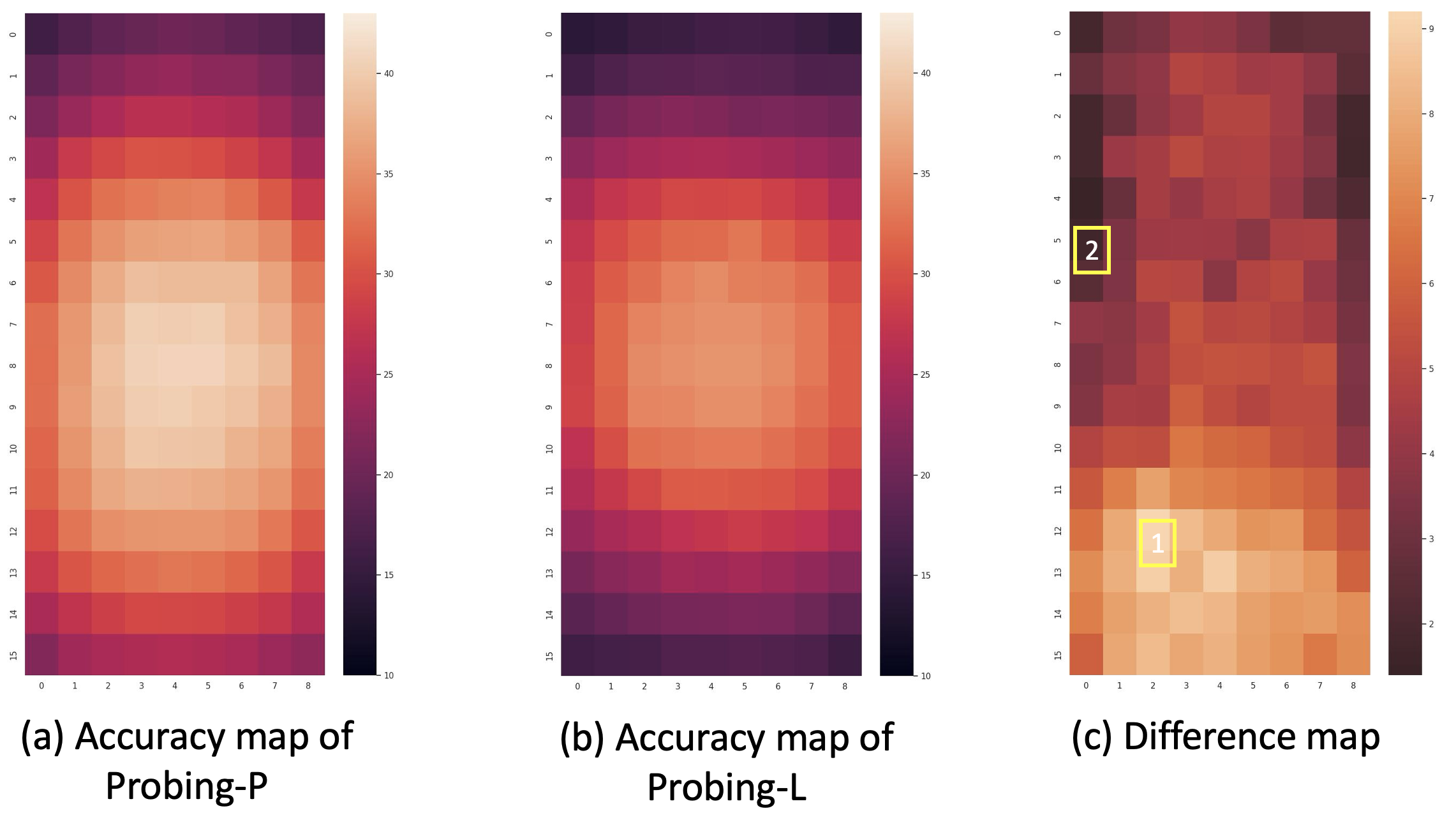}
\end{center}
\vspace{-4mm}
% \caption{Visualization results on S100. Our framework is able to deal with . Why illustrating the 1 and 2 crops in this figure rather than Fig.1}
\caption{
The heatmaps of evaluating the Probing-P (a) and Probing-L (b) at different spatial locations on the validation set of S100-PM. (c) shows the accuracy differences between Probing-P and Probing-L.}
\label{fig:visualize_pm}
\vspace{-2mm}
\end{figure}

Using Probing-P and Probing-L, we compose an accuracy map  of size $16\times9$ from the accuracies obtained from the different evaluation positions on the S100-PM validation set as shown in Figure~\ref{fig:visualize_pm} (a) and Figure~\ref{fig:visualize_pm} (b).
%\Mingfei{We assume that the inference accuracy of a fixed frame crop indicates the relativity between training videos and testing videos in the particular position.}
We further compute the difference between the two heat maps in Figure~\ref{fig:visualize_pm} (a) and (b) and obtain the difference map as in Figure~\ref{fig:visualize_pm} (c).
\Mingfei{Here, the difference value in each position indicates the gap of recognition abilities of the same model trained on landscape mode videos and portrait mode videos, respectively.}
If a value on the different map is greater than 0, it indicates that Probing-P achieves higher accuracy than Probing-L.
%It indicates that PM videos are more informative to recognise videos cropped at the position and vice versa.
% For example, as outlined by yellow boxes in Figure~\ref{fig:visualize_pm} (c), mark 1 indicates a nearly identical ability of recognition for LM and PM videos.
% Similarly, mark 2 indicates PM videos are much more informative at that position with an accuracy gap greater than 8 points.
For example, as outlined by the yellow boxes in Figure~\ref{fig:visualize_pm} (c), mark 1 indicates the model trained with PM videos is stronger to recognize the video categories at this location, while mark 2 indicates models trained by PM and LM videos have similar performance at this location.
In general, it can be inferred from the brighter areas in Figure~\ref{fig:visualize_pm} (a) that informative areas in PM videos are more densely concentrated at the middle to lower half of the video. %On the other hand, 
%\Heng{In other words}, 
It can also be inferred from Figure~\ref{fig:visualize_pm} (c) that the bottom part of the PM videos contains specific domain knowledge that does not exist in the LM videos, leading to bad performance of models trained on LM videos in this region.  

\begin{figure}[t]
\begin{center}
   \includegraphics[width=1.0\linewidth]{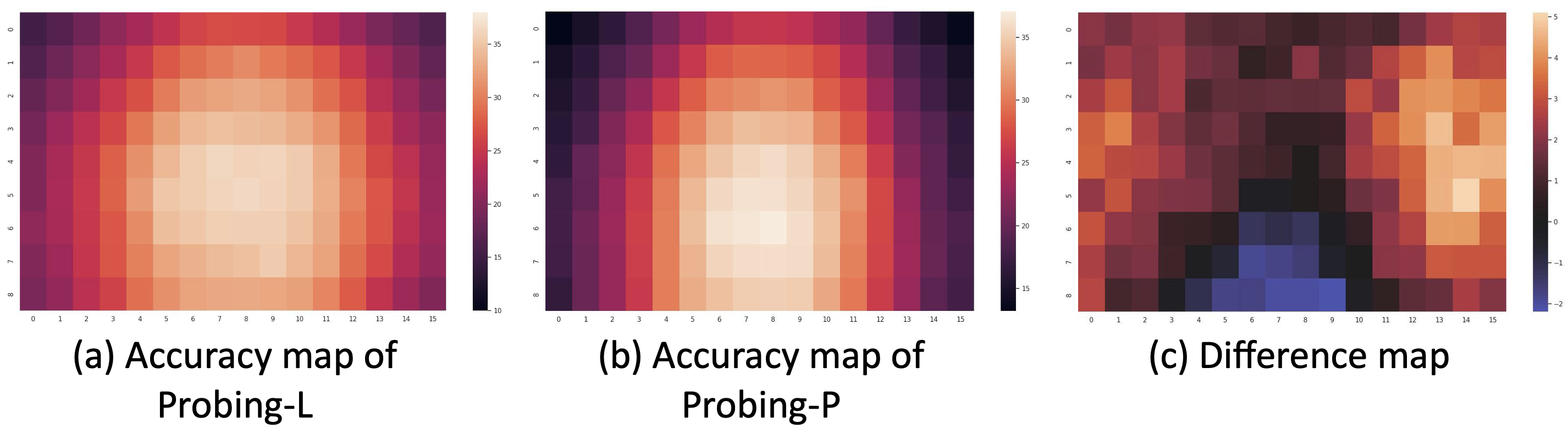}
\end{center}
\vspace{-3mm}
\caption{The heatmaps of evaluating the Probing-L (a) and Probing-P (b) at different spatial locations on the validation set of S100-LM. (c) shows the accuracy difference between Probing-L and Probing-P.}
\label{fig:visualize_lm}
\vspace{-4mm}
\end{figure}

Similarly, we show the accuracy heat maps of the Probing-L and Probing-P evaluated on the LM videos in Figure~\ref{fig:visualize_lm} (a) (b), with the difference of the two heat maps shown in  Figure~\ref{fig:visualize_lm} (c).
%accuracy heat maps created by LM videos and PM videos are shown in (a) and (b) respectively, and the difference heat map is shown in (c).
%Interestingly and differently, the informative areas of LM videos, shown in warm colours in Figure~\ref{fig:visualize_lm} (c), are distributed in the shape of dumbbell.
It can be seen that the informative areas in LM videos are in the center part of the video, and the left and right sides on the video frame contain specific domain knowledge that cannot be learned from PM videos. For example, some actions with a wide background in LM videos may not have similar visual cues in the PM videos.

% As shown in line 1-2 of \cref{ab_s100}, X3D trained with PM videos outperforms model trained with LM videos by a large margin of ~8(\%) on PM validation videos.
% When evaluated with LM validation videos, X3D achieves comparable performance either trained with PM videos or LM videos.
% This indicates that PM< videos are necessary to achieve satisfying performance on PM videos.
% Moreover, models trained with PM videos can be easily applied to LM videos which achieves comparable performance, which shows that portrait mode videos have different data prior to traditional landscape mode videos.

\section{Comparison of data preprocessing recipes}
% As videos in different display modes indicate different data prior,
% In this subsection, extensive experiments are performed under different data preprocessing recipes with various datasets to find an effective practice for portrait mode videos.
Effective data preprocessing is essential for achieving high performance in video classification tasks. 
% However, handling videos in portrait mode presents unique challenges due to their aspect ratio being different from the conventional landscape mode.
In this section, we investigate the impact of different data preprocessing strategies on the performance of portrait mode video recognition.
We hypothesize that videos in different aspect ratios
%orientations 
may require different crop resolutions for optimal performance. 
To test this hypothesis, we perform extensive experiments on various portrait mode video datasets, using different crop resolutions and data augmentation techniques. 
Through our experiments, \Heng{we identify the best recipes for portrait mode videos when using CNN or transformer models, which are different from that of landscape mode videos.}

%find that different optimal crop resolutions exist 
%for CNN models and transformer models.
% These findings provide valuable insights into effective data preprocessing practices for handling portrait mode videos in video classification tasks.

\subsection{Resizing and area sampling}
% In the current video recognition pipeline, the first step of processing a video is to resize and crop the input frames to meet the computational resource and inference cost requirements. 
Resizing and cropping are critical steps in the data preprocessing pipeline for video recognition, as they allow videos to be processed efficiently and \Heng{are also important ways of data augmentation.}
%while still maintaining key information for accurate classification. 
% Two popular methods for resizing and cropping input frames are the inception-style method \cite{} and the shorter-side resizing method.
Different models in various architectures adopt different strategies. 
The two popular strategies are the Inception-style method \cite{szegedy2015going,feichtenhofer2019slowfast,fan2021mvit,li2022uniformer,touvron2021training,liu2021videoswin}, 
% which randomly scales and crops frames, 
and the shorter-side resizing method \cite{simonyan2014very}.
% which resizes frames along the shorter side and crops in fixed resolution, \eg, 224$\times$224. 
In this subsection, we will explore these two methods in more detail and investigate their effectiveness for portrait mode video recognition.
% However, the frame resizing and crop strategy needs to be revisited for portrait mode video recognition, as this type of video only contains frames in portrait orientation. In this section, we experiment with different models and observe different optimal crop strategies for CNN models and transformer models, which can provide insights into selecting proper crops when adopting different architectures.

The shorter-side resizing method is widely used in video recognition methods \cite{carreira2017quo,wang2018non,tran2019video,feichtenhofer2020x3d,wang2020video,bertasius2021space,neimark2021video,chen2021deep,zhu2022uni,wu2020multigrid,sun2021dynamic,li2020smallbignet,wang2021tdn,wang2018videos,xiao2022learning,yang2020temporal}.
It involves resizing the video frames so that the shorter side of the frame is set to a length that is fixed \cite{chen2021deep,xiao2022learning} or randomly sampled within a range \cite{carreira2017quo,wang2018non,tran2019video,feichtenhofer2020x3d,wang2020video,bertasius2021space,neimark2021video,zhu2022uni,wu2020multigrid,sun2021dynamic,li2020smallbignet,wang2021tdn,wang2018videos,yang2020temporal},
while the longer side is scaled proportionally.
Then the frames are centre-cropped to a square shape, typically 224$\times$224 and passed into the model.
This approach ensures that the input frames have a consistent aspect ratio and are cropped without distortion.
In contrast,
% to the fixed sampling area and fixed crop aspect ratio, 
the Inception-style method augments the shorter-side resizing method with two additional random sampling steps.
% The inception-syle method was originally proposed in Inception v1\cite{} and has since become a standard preprocessing step in many video recognition pipelines \cite{}.
The first one is to sample a target pixel number from the whole-size video frame by the random ratio between 8\% and 100\%. 
Then, it randomly samples an aspect ratio between 3/4 and 4/3 and reshapes the crop area accordingly. 
Finally, it crops the frames at a random position and resizes them to a fixed resolution in squares (\eg, 224$\times$224) without keeping the aspect ratio.
This approach can sample a diverse set of inputs and is designed to adapt the model to videos in different sizes.

\begin{table}[t]
    \small
    \begin{center}
        \small
\setlength{\tabcolsep}{1.2mm}
\begin{tabular}
{@{} l | l | c c | c @{}}
% \toprule
% \multirow{2}{*}{Model} & \multirow{2}{*}{Data} & \multicolumn{2}{c}{Recipes} \\
% \arrayrulecolor{white}\cline{3-4}
% \arrayrulecolor{black}\cline{3-4}
% \arrayrulecolor{black}\cline{3-4}
% \arrayrulecolor{white}\cline{3-4}
%  & & Inception-style & Shorter-side \\
% \arrayrulecolor{white}\hline
% \arrayrulecolor{black}\hline
% \arrayrulecolor{white}\hline
\multicolumn{1}{c|}{Model} & \multicolumn{1}{c|}{Data} & Incep. & Short. & GFLOPs\x views \\
\shline
\multirow{2}{*}{X3D-M\cite{feichtenhofer2020x3d}} & S100-PM & \textbf{54.2} & 52.0 & 4.9$\times$3$\times$10\\
% \arrayrulecolor{white}\cline{2-4}
% \arrayrulecolor{white}\cline{2-4}
% \arrayrulecolor{black}\cline{2-4}
% \arrayrulecolor{white}\cline{2-4}
% \arrayrulecolor{white}\cline{2-4}
                          & 3Massiv & \textbf{53.7} & 52.6 & 4.9$\times$3$\times$10\\
% \arrayrulecolor{white}\cline{2-4}
% \arrayrulecolor{white}\cline{2-4}
% \arrayrulecolor{black}\cline{2-4}
% \arrayrulecolor{white}\cline{2-4}
% \arrayrulecolor{white}\cline{2-4}
                       & \abbrdataset & \textbf{61.7} & 61.2 & 4.9$\times$3$\times$10\\
\arrayrulecolor{white}\hline
\arrayrulecolor{black}\hline
\arrayrulecolor{white}\hline
\multirow{2}{*}{Uniformer-S\cite{li2022uniformer}} & S100-PM & 39.7 & \textbf{42.0} & 41.8$\times$1$\times$4\\
% \arrayrulecolor{white}\cline{2-4}
% \arrayrulecolor{white}\cline{2-4}
% \arrayrulecolor{black}\cline{2-4}
% \arrayrulecolor{white}\cline{2-4}
% \arrayrulecolor{white}\cline{2-4}
                       & 3Massiv & 42.8 & \textbf{43.6} & 41.8$\times$1$\times$4 \\
% \arrayrulecolor{white}\cline{2-4}
% \arrayrulecolor{white}\cline{2-4}
% \arrayrulecolor{black}\cline{2-4}
% \arrayrulecolor{white}\cline{2-4}
% \arrayrulecolor{white}\cline{2-4}
                       & \abbrdataset & 50.2 & \textbf{50.4} & 64.0$\times$1$\times$5 \\
\arrayrulecolor{white}\hline
\arrayrulecolor{black}\hline
\arrayrulecolor{white}\hline
\multirow{2}{*}{MViTv2-S\cite{li2022mvitv2}} & S100-PM & 36.9 & \textbf{41.0} & 64.0$\times$1$\times$5 \\
% \arrayrulecolor{white}\cline{2-4}
% \arrayrulecolor{white}\cline{2-4}
% \arrayrulecolor{black}\cline{2-4}
% \arrayrulecolor{white}\cline{2-4}
% \arrayrulecolor{white}\cline{2-4}
                          & 3Massiv & 50.4 & \textbf{52.1} & 64.0$\times$1$\times$5 \\
% \arrayrulecolor{white}\cline{2-4}
% \arrayrulecolor{white}\cline{2-4}
% \arrayrulecolor{black}\cline{2-4}
% \arrayrulecolor{white}\cline{2-4}
% \arrayrulecolor{white}\cline{2-4}
                       % & \abbrdataset & 80.70 & 80.68 \\
                       & \abbrdataset & 61.7 & \textbf{62.0} & 64.0$\times$1$\times$5 \\
% \arrayrulecolor{white}\hline
% \arrayrulecolor{black}\hline
% \arrayrulecolor{white}\hline

\end{tabular}
    \end{center}
    \vspace{-5mm}
    \caption{Comparison of top-1 accuracy (\%) of different resizing and area sampling strategies for portrait mode videos, \ie, inception style (Incep.) and shorter-side style(Short.). Views during inference are shown by
the multiplication of \# of spatial crops and \# of temporal views.}
    \label{tab:ab_res} 
    % \vspace{-1mm}
\end{table}

\begin{table}[t]
    \small
    \begin{center}
        \small
\setlength{\tabcolsep}{1.2mm}
\begin{tabular}
{@{} l | l | c c c @{}}
% \toprule
% Data & Model & 224$\times$224 & 256$\times$192 & 288$\times$192\\
% \arrayrulecolor{white}\cline{3-5}
% \arrayrulecolor{black}\cline{3-5}
% \arrayrulecolor{black}\cline{3-5}
% \arrayrulecolor{white}\cline{3-5}
% \arrayrulecolor{white}\hline
% \arrayrulecolor{black}\hline
% \arrayrulecolor{white}\hline
% \multirow{2}{*}{S100-PM} & Uniformer-S & 42.18 & 43.30 & 45.55 \\
% \arrayrulecolor{white}\cline{2-5}
% \arrayrulecolor{white}\cline{2-5}
% \arrayrulecolor{black}\cline{2-5}
% \arrayrulecolor{white}\cline{2-5}
% \arrayrulecolor{white}\cline{2-5}
%                        & MViTv2-S & 41.04 & 40.03 & 45.49 \\
% \arrayrulecolor{white}\hline
% \arrayrulecolor{black}\hline
% \arrayrulecolor{white}\hline
% \multirow{2}{*}{3Massiv} & Uniformer-S & 44.75 & 45.24 & 46.87 \\
% \arrayrulecolor{white}\cline{2-5}
% \arrayrulecolor{white}\cline{2-5}
% \arrayrulecolor{black}\cline{2-5}
% \arrayrulecolor{white}\cline{2-5}
% \arrayrulecolor{white}\cline{2-5}
%                           & MViTv2-S & 52.08 & 52.30 & 53.78 \\
% \arrayrulecolor{white}\hline
% \arrayrulecolor{black}\hline
% \arrayrulecolor{white}\hline

\multirow{2}{*}{~~~~~~~~Model} & \multirow{2}{*}{~~~~Data} & \multicolumn{3}{c}{Training crops}\\
% \multirow{2}{*}{Model} & \multirow{2}{*}{Data} & 224$\times$224 & 256$\times$192 &  288$\times$192 \\
  % &  & xxG & xxG & xxG \\
% \arrayrulecolor{white}\cline{3-5}
% \arrayrulecolor{black}\cline{3-5}
% \arrayrulecolor{black}\cline{3-5}
% \arrayrulecolor{white}\cline{3-5}
% \arrayrulecolor{white}\cline{3-5}
 & & 224$\times$224 & 256$\times$192 & 288$\times$192 \\
% \arrayrulecolor{white}\hline
% \arrayrulecolor{black}\hline
% \arrayrulecolor{white}\hline
\shline
\multirow{3}{*}{X3D-M\cite{feichtenhofer2020x3d}} & S100-PM & \textbf{52.0} & 51.6 & 50.8 \\
% \arrayrulecolor{white}\cline{2-5}
% \arrayrulecolor{white}\cline{2-5}
% \arrayrulecolor{black}\cline{2-5}
% \arrayrulecolor{white}\cline{2-5}
% \arrayrulecolor{white}\cline{2-5}
                          & 3Massiv & \textbf{52.6} & 52.5 & 50.8 \\
% \arrayrulecolor{white}\cline{2-5}
% \arrayrulecolor{white}\cline{2-5}
% \arrayrulecolor{black}\cline{2-5}
% \arrayrulecolor{white}\cline{2-5}
% \arrayrulecolor{white}\cline{2-5}
                          & \abbrdataset & \textbf{61.2} & 61.0 & 60.8 \\
\arrayrulecolor{white}\hline
\arrayrulecolor{black}\hline
\arrayrulecolor{white}\hline
\multirow{3}{*}{Uniformer-S\cite{li2022uniformer}} & S100-PM & 42.0 & 43.3 & \textbf{45.4} \\
% \arrayrulecolor{white}\cline{2-5}
% \arrayrulecolor{white}\cline{2-5}
% \arrayrulecolor{black}\cline{2-5}
% \arrayrulecolor{white}\cline{2-5}
% \arrayrulecolor{white}\cline{2-5}
                       & 3Massiv & 43.6 & 44.6 & \textbf{45.8} \\
% \arrayrulecolor{white}\cline{2-5}
% \arrayrulecolor{white}\cline{2-5}
% \arrayrulecolor{black}\cline{2-5}
% \arrayrulecolor{white}\cline{2-5}
% \arrayrulecolor{white}\cline{2-5}
                       & \abbrdataset & 50.4 & 50.8 & \textbf{51.6} \\
\arrayrulecolor{white}\hline
\arrayrulecolor{black}\hline
\arrayrulecolor{white}\hline
\multirow{3}{*}{MViTv2-S\cite{li2022mvitv2}} & S100-PM & 41.0 & 40.0 & \textbf{45.5} \\
% \arrayrulecolor{white}\cline{2-5}
% \arrayrulecolor{white}\cline{2-5}
% \arrayrulecolor{black}\cline{2-5}
% \arrayrulecolor{white}\cline{2-5}
% \arrayrulecolor{white}\cline{2-5}
                          & 3Massiv & 52.1 & 52.3 & \textbf{53.8} \\
% \arrayrulecolor{white}\cline{2-5}
% \arrayrulecolor{white}\cline{2-5}
% \arrayrulecolor{black}\cline{2-5}
% \arrayrulecolor{white}\cline{2-5}
% \arrayrulecolor{white}\cline{2-5}
                          % & \abbrdataset & 80.68 & 80.55 & \textbf{80.84} \\
                          & \abbrdataset &62.0 & 61.4 & \textbf{62.8} \\
% \arrayrulecolor{white}\hline
% \arrayrulecolor{black}\hline
% \arrayrulecolor{white}\hline

\end{tabular}
    \end{center}
    \vspace{-3mm}
    \caption{Top-1 accuracy (\%) of different training crop resolutions. The models are always tested with the same square crops in \textbf{224$\times$224} to ensure the same inference cost across different training crop resolutions.  }
    % trained with different crop resolutions and
    \label{tab:ab_shape_attn}
    \vspace{-2mm}
\end{table}

% Apart from existing data setups, portrait mode video recognition only deals with videos in vertical orientations.
% Different data priors inspire us to dive into the pre-processing pipeline of video recognition for a practical recipe for video recognition in portrait mode.
We carry out extensive experiments on models of different architectures with the two resizing strategies in Table~\ref{tab:ab_res}. 
To alleviate the bias introduced by mixed-orientation data, the models are trained from scratch and we keep any other training setup identical to their original papers, except for learning hyper-parameters, such as batch size and learning rate. 
During inference, identical augmentation and sampling methods are adopted for different recipes.
We guide the readers to supplemental materials for more details. 

As shown in Table~\ref{tab:ab_res}, each model is evaluated on three different portrait mode video benchmarks.
For the CNN-based model, \ie, X3D-M \cite{feichtenhofer2020x3d}, the random scaling strategy from the Inception-style method brings an improvement of 2.2\% (54.2\% vs. 52.0\%) on S100-PM \cite{kinetics700}, 1.1\% (53.7\% vs. 52.6\%) on 3Massiv \cite{gupta20223massiv} and 0.5\% on \abbrdataset.
Differently, as for the transformer-based models, \ie, Uniformer-S \cite{li2022uniformer} and MViTv2-S \cite{li2022mvitv2}, randomly scaled input crops bring down the accuracy by a large margin.
For example, the random scaling reduces the performance of Uniformer-S by 2.3\% (42.0\% vs. 39.7\%) on S100-PM, 0.8\% (43.6\% vs. 42.8\%) on 3Massiv and 1.3\% (72.1\% vs. 70.8\%) on \abbrdataset.
% A similar phenomenon can also be observed on MViTv2-S with performance downgrades of 0.83\% to 4.1\% on different benchmarks.
MViTv2-S also shows performance drops from 0.3\% to 4.1\% across benchmarks.
% It is interesting to see that better recipes for the various models are different from their practices on hybrid orientation benchmarks, \ie, Kinetics \cite{kay2017kinetics}.
This suggests that optimal strategies diverge from those used in mixed orientation benchmarks like Kinetics\cite{kay2017kinetics}.
% Interestingly, MViTv2-S also shows performance drops from 0.83% to 4.1% across benchmarks. This suggests that optimal strategies for different models diverge from those used in mixed orientation benchmarks like Kinetics.
 \begin{table}[t]
    \small
    \begin{center}
        \small
\setlength{\tabcolsep}{2.4mm}
\begin{tabular}
{@{} l | l | c c @{}}
% \toprule
% Data & Model & 224$\times$224 & 256$\times$192 & 288$\times$192\\
% \arrayrulecolor{white}\cline{3-5}
% \arrayrulecolor{black}\cline{3-5}
% \arrayrulecolor{black}\cline{3-5}
% \arrayrulecolor{white}\cline{3-5}
% \arrayrulecolor{white}\hline
% \arrayrulecolor{black}\hline
% \arrayrulecolor{white}\hline
% \multirow{2}{*}{S100-PM} & Uniformer-S & 42.18 & 43.30 & 45.55 \\
% \arrayrulecolor{white}\cline{2-5}
% \arrayrulecolor{white}\cline{2-5}
% \arrayrulecolor{black}\cline{2-5}
% \arrayrulecolor{white}\cline{2-5}
% \arrayrulecolor{white}\cline{2-5}
%                        & MViTv2-S & 41.04 & 40.03 & 45.49 \\
% \arrayrulecolor{white}\hline
% \arrayrulecolor{black}\hline
% \arrayrulecolor{white}\hline
% \multirow{2}{*}{3Massiv} & Uniformer-S & 44.75 & 45.24 & 46.87 \\
% \arrayrulecolor{white}\cline{2-5}
% \arrayrulecolor{white}\cline{2-5}
% \arrayrulecolor{black}\cline{2-5}
% \arrayrulecolor{white}\cline{2-5}
% \arrayrulecolor{white}\cline{2-5}
%                           & MViTv2-S & 52.08 & 52.30 & 53.78 \\
% \arrayrulecolor{white}\hline
% \arrayrulecolor{black}\hline
% \arrayrulecolor{white}\hline

\multirow{2}{*}{~~~~~~Model} & \multirow{2}{*}{~~~~Data} & \multicolumn{2}{c}{Testing crops} \\
% \arrayrulecolor{white}\cline{3-4}
% \arrayrulecolor{black}\cline{3-4}
% \arrayrulecolor{black}\cline{3-4}
% \arrayrulecolor{white}\cline{3-4}
 & & 256$\times$192 &  288$\times$192 \\
% \arrayrulecolor{white}\hline
% \arrayrulecolor{black}\hline
% \arrayrulecolor{white}\hline
\shline
\multirow{3}{*}{X3D-M\cite{feichtenhofer2020x3d}} & S100-PM & 51.4$_{0.2\downarrow}$ & 50.4$_{0.4\downarrow}$ \\
% \arrayrulecolor{white}\cline{2-4}
% \arrayrulecolor{white}\cline{2-4}
% \arrayrulecolor{black}\cline{2-4}
% \arrayrulecolor{white}\cline{2-4}
% \arrayrulecolor{white}\cline{2-4}
                          & 3Massiv & 52.6$_{0.1\uparrow}$ & 52.0$_{1.2\uparrow}$\\
% \arrayrulecolor{white}\cline{2-4}
% \arrayrulecolor{white}\cline{2-4}
% \arrayrulecolor{black}\cline{2-4}
% \arrayrulecolor{white}\cline{2-4}
% \arrayrulecolor{white}\cline{2-4}
                          & \abbrdataset & 62.9$_{1.9\mathbf{\uparrow}}$ & 63.1$_{2.3\uparrow}$ \\
% \arrayrulecolor{white}\hline
% \arrayrulecolor{black}\hline
% \arrayrulecolor{white}\hline
\hline
\multirow{3}{*}{Uniformer-S\cite{li2022uniformer}} & S100-PM & 44.4$_{1.1\mathbf{\uparrow}}$ & 46.5$_{1.1\mathbf{\uparrow}}$ \\
% \arrayrulecolor{white}\cline{2-4}
% \arrayrulecolor{white}\cline{2-4}
% \arrayrulecolor{black}\cline{2-4}
% \arrayrulecolor{white}\cline{2-4}
% \arrayrulecolor{white}\cline{2-4}
                       & 3Massiv & 45.7$_{1.1\mathbf{\uparrow}}$ & 47.3$_{1.5\mathbf{\uparrow}}$ \\
% \arrayrulecolor{white}\cline{2-4}
% \arrayrulecolor{white}\cline{2-4}
% \arrayrulecolor{black}\cline{2-4}
% \arrayrulecolor{white}\cline{2-4}
% \arrayrulecolor{white}\cline{2-4}
                       & \abbrdataset & 51.9$_{1.1\mathbf{\uparrow}}$ & 53.3$_{1.7\mathbf{\uparrow}}$ \\
% \arrayrulecolor{white}\hline
% \arrayrulecolor{black}\hline
% \arrayrulecolor{white}\hline
\hline
\multirow{3}{*}{MViTv2-S\cite{li2022mvitv2}} & S100-PM & 39.8$_{0.2\downarrow}$ & 46.8$_{1.30\mathbf{\uparrow}}$ \\
% \arrayrulecolor{white}\cline{2-4}
% \arrayrulecolor{white}\cline{2-4}
% \arrayrulecolor{black}\cline{2-4}
% \arrayrulecolor{white}\cline{2-4}
% \arrayrulecolor{white}\cline{2-4}
                          & 3Massiv & 52.7$_{0.4\mathbf{\uparrow}}$ & 54.8$_{1.0\mathbf{\uparrow}}$ \\
% \arrayrulecolor{white}\cline{2-4}
% \arrayrulecolor{white}\cline{2-4}
% \arrayrulecolor{black}\cline{2-4}
% \arrayrulecolor{white}\cline{2-4}
% \arrayrulecolor{white}\cline{2-4}
                          % & \abbrdataset & 80.68 & 80.55 & \textbf{80.84} \\
                          & \abbrdataset & 62.1$_{0.7\uparrow}$ & 63.7$_{0.9\mathbf{\uparrow}}$ \\
% \arrayrulecolor{white}\hline
% \arrayrulecolor{black}\hline
% \arrayrulecolor{white}\hline

\end{tabular}
    \end{center}
    \vspace{-4mm}
    \caption{Top-1 accuracy (\%) of using the same resolution for both training and testing crops.
    %on test set for different input crop shapes. Models are tested with crops in the same resolution as the training crop. 
    We also report the performance difference compared with using 224$\times$224 testing crops from the first column of Table~\ref{tab:ab_shape_attn}, where $\uparrow$ means higher result.
    % than tested with 224$\times$224.  
    %The two columns correspond to the last two columns in Table~\ref{tab:ab_shape_attn}.
% The models are tested with crops in 224$\times$224 squares, and rectangles in the same shapes as the training resolution.
    }
    \label{tab:ab_shape_attn_rect}
    % \vspace{-1mm}
\end{table}

It may be hard to determine the cause of the interesting phenomenon, but we can make a reasonable assumption that it is due to the different data priors in portrait mode only video benchmarks, such as S100-PM and \abbrdataset. 
With portrait mode videos, the object and its movement are typically limited to a vertical space, which may result in unique visual patterns that are not present in hybrid orientation benchmarks, such as Kinetics. 
% Further research is needed to better understand the impact of data prior on video recognition performance. 
While the cause requires further investigation, these results suggest that there may be unique characteristics of portrait mode videos that require specialized recognition methods.

% We first ablate on the frame resizing and crop area sampling strategies.
% For attention-based models, \ie, Uniformer and MViT v2, Inception-style recipe is adopted by default (randomly resize the input area between a [min, max] range, scale of [0.08, 100], and jitter aspect ratio between 3/4 and 4/3).
% The region of classification, \ie, the cropped input is stretched and interpolated to a square, without aspect ratio of the content kept.
% We compare the default recipe with the shorter-side resizing recipe, in which the frame is resized along the shorter side to a fixed length and a square input is directly cropped as region of classification, with aspect ratio of the content kept.

% As shown in Table~\ref{tab:ab_res}, shorter-side recipe brings an improvement of ~2 points without additional computation introduced.
% We suspect that the improvement is sourced from the different data prior of portrait mode videos.
% As the actor and action content is naturally distributed vertically, the portrait mode videos can capture as much informative area as possible, which is different from landscape mode videos which displayed horizontally capturing more background and surroundings.
% As such, the informative content in portrait mode videos is more likely to suffer from severe deformation and distortion, brought by the jittering aspect ratio and stretching interpolation of input crop.

\begin{table*}[t]
\centering
% \captionsetup{captionskip=2pt}
% minipage for the first sub-table
\begin{minipage}{.49\linewidth} % Adjust the width as needed
  \centering
  \small
\setlength{\tabcolsep}{2.0mm}
\begin{tabular}
{@{} l | c | c | c @{}}
% \toprule

\multicolumn{1}{c|}{Data} & Model & \# of Frames & Top1-Acc.\\
\shline
% \arrayrulecolor{white}\cline{1-5}
% \arrayrulecolor{black}\cline{1-5}
% \arrayrulecolor{black}\cline{1-5}
% \arrayrulecolor{white}\cline{1-5}
% \arrayrulecolor{white}\hline
% \arrayrulecolor{black}\hline
% \arrayrulecolor{white}\hline
\multirow{2}{*}{K400~\cite{kay2017kinetics}} & Uniformer-frames & 16$\times$4 & 72.1 \\
% \arrayrulecolor{white}\cline{2-4}
% \arrayrulecolor{white}\cline{2-4}
% \arrayrulecolor{black}\cline{2-4}
% \arrayrulecolor{white}\cline{2-4}
% \arrayrulecolor{white}\cline{2-4}
                       & Uniformer~\cite{li2022uniformer} & 16$\times$4 & 76.6$_{4.5\uparrow}$ \\
\hline
% \arrayrulecolor{white}\hline
% \arrayrulecolor{black}\hline
% \arrayrulecolor{white}\hline
% \multirow{2}{*}{Ssv2} & Uniformer-TSN & 16$\times$4 & \textcolor{red}{xx} \\
% \arrayrulecolor{white}\cline{2-4}
% \arrayrulecolor{white}\cline{2-4}
% \arrayrulecolor{black}\cline{2-4}
% \arrayrulecolor{white}\cline{2-4}
% \arrayrulecolor{white}\cline{2-4}
%                        & Uniformer & 16$\times$4 & \textcolor{red}{xx} \\
% \arrayrulecolor{white}\hline
% \arrayrulecolor{black}\hline
% \arrayrulecolor{white}\hline
\multirow{2}{*}{3Massiv~\cite{gupta20223massiv}} & Uniformer-frames & 16$\times$4 & 41.9 \\
% \arrayrulecolor{white}\cline{2-4}
% \arrayrulecolor{white}\cline{2-4}
% \arrayrulecolor{black}\cline{2-4}
% \arrayrulecolor{white}\cline{2-4}
% \arrayrulecolor{white}\cline{2-4}
                       & Uniformer~\cite{li2022uniformer} & 16$\times$4 & 42.8$_{0.9\uparrow}$ \\
% \arrayrulecolor{white}\hline
% \arrayrulecolor{black}\hline
% \arrayrulecolor{white}\hline
\hline
\multirow{2}{*}{\abbrdataset} & Uniformer-frames & 16$\times$4 & 45.7 \\
% \arrayrulecolor{white}\cline{2-4}
% \arrayrulecolor{white}\cline{2-4}
% \arrayrulecolor{black}\cline{2-4}
% \arrayrulecolor{white}\cline{2-4}
% \arrayrulecolor{white}\cline{2-4}
                       & Uniformer~\cite{li2022uniformer} & 16$\times$4 & 50.3$_{4.6\uparrow}$ \\
% \arrayrulecolor{white}\hline
% \arrayrulecolor{black}\hline
% \arrayrulecolor{white}\hline

\end{tabular}
  \vspace{-1mm}
  \captionof{table}{\textbf{Temporal information importance}: Effect of utilizing temporal information for video recognition on different benchmarks.}
  \label{tab:ab_temporal_info}
\end{minipage}
\hspace{5mm} % Space between the sub-tables
% minipage for the second sub-table
\begin{minipage}{.38\linewidth} % Adjust the width as needed
  \centering
  \small
\setlength{\tabcolsep}{3.0mm}
\begin{tabular}
{@{} l | c | c @{}}
% \toprule

\multicolumn{1}{c|}{Data} & Modality & Top1-Acc.\\
\shline
\multirow{3}{*}{3Massiv~\cite{gupta20223massiv}} & Visual & 52.7 \\
                       & Audio & 31.6 \\
% \arrayrulecolor{white}\cline{2-3}
% \arrayrulecolor{black}\cline{2-3}
% \arrayrulecolor{white}\cline{2-3}
                       & Visual+Audio & 54.9 \\
\arrayrulecolor{black}\hline{}
\multirow{3}{*}{\abbrdataset} & Visual & 54.6 \\
                       & Audio & 15.2 \\
% \arrayrulecolor{white}\cline{2-3}
% \arrayrulecolor{black}\cline{2-3}
% \arrayrulecolor{white}\cline{2-3}
                       & Visual+Audio & 57.0 \\

\end{tabular}
  \vspace{-1mm}
  \captionof{table}{\textbf{Audio importance}: Comparison of different modalities with offline feature embeddings.}
  \label{tab:ab_audio}
\end{minipage}
% \vspace{0.5em}
% \caption{Ablations on the Fast pathway design on \textbf{Kinetics-400}. We show top-1 and top-5 classification accuracy (\%), as well as computational complexity measured in GFLOPs (floating-point operations, in \# of multiply-adds \x $10^9$) for a single clip input of spatial size 256$^2$. Inference-time computational cost is proportional to this, as a fixed number of 30 of views is used. Backbone: 4\x 16, R-50.}
% \label{tab:ablations}
\vspace{-0.8em}
\end{table*}

\subsection{Shape of frame crop}

 In this subsection, we explore the impact of different crop strategies on model performance in portrait mode video recognition. Specifically, we investigate the performance of models trained and tested on crops of varying sizes and aspect ratios.
 
 Traditional methods typically use square frame crops to ensure even coverage of object and movement in both vertical and horizontal directions. 
 % However, we propose that this approach may not be optimal for portrait mode videos.
 However, we argue that this approach may not be optimal for portrait mode videos, which typically contain object and movement information in vertical directions. 
 Cropping the frames into squares could potentially result in a loss of critical information and more background noise. 
 As shown in Figure~\ref{fig:visualize_pm}, portrait mode videos possess more informative content distributed vertically, and cropping into squares may not effectively capture this information.

 % As described in Section~\ref{sec:pm_lm_vis}, portrait mode videos possess more informative content distributed vertically compared to landscape mode videos, highlighted by the warm colours in Figure~\ref{fig:visualize_pm}.
 % Whereas, landscape mode videos possess more informative content distributed horizontally compared to portrait mode videos, highlighted by the warm colours in Figure~\ref{fig:visualize_lm}.
 % Square frame crops ensure the informative video content in both orientations can be covered.
 % However, cropping portrait mode videos into squares may not effectively capture the vertically distributed critical information.

 To comply with the unique information distributive characteristics, we propose to crop the areas in vertical rectangles and input them directly into models without distortion. 
 We experiment with crops in different aspect ratios and in similar pixel numbers to the square input, \ie, 256$\times$192 and 288$\times$192, in order to fairly compare the models under different input resolutions.
 %op strategy for portrait mode video recognition.
 With input shape changed, we only modify the last global pooling layer.
 We keep any other training details identical to the setup using square inputs.
 
 As shown in Table~\ref{tab:ab_shape_attn}, we train models with different input crops on portrait mode video benchmarks and test with square crops, \ie, 224$\times$224 to ensure identical inference cost.
 It is thrilled to see that increase in aspect ratio introduces continuing performance improvement for transformer-based models, \ie, Uniformer-S and MViTv2-S.
 We also observe that change in aspect ratio degrades the performance of X3D-M, showing different behaviour to transformer-based models.
 %The cause may be the convolution networks natures.
 The potential reason could be due to the fixed square receptive field of convolution networks regardless of the input resolutions, which is not compatible with the elongated image shape.
% \Mingfei{The possible reason X3D-M observe a performance decrease}

% \begin{table*}[t]
% \centering
% \captionsetup{captionskip=2pt}
% % minipage for the first sub-table
% \begin{minipage}{.5\linewidth} % Adjust the width as needed
%   \centering
%   \input{tables/temporal_info}
%   \captionof{table}{\textbf{Temporal information importance} Effect of utilizing temporal information for video recognition on different benchmarks.}
%   \label{tab:ab_temporal_info}
% \end{minipage}
% \hspace{5mm} % Space between the sub-tables
% % minipage for the second sub-table
% \begin{minipage}{.38\linewidth} % Adjust the width as needed
%   \centering
%   \input{tables/2stage_3massiv}
%   \captionof{table}{\textbf{Audio importance}: Comparison of different modalities with offline feature embeddings.}
%   \label{tab:ab_audio_3massiv}
% \end{minipage}
% \vspace{0.5em}
% % \caption{Ablations on the Fast pathway design on \textbf{Kinetics-400}. We show top-1 and top-5 classification accuracy (\%), as well as computational complexity measured in GFLOPs (floating-point operations, in \# of multiply-adds \x $10^9$) for a single clip input of spatial size 256$^2$. Inference-time computational cost is proportional to this, as a fixed number of 30 of views is used. Backbone: 4\x 16, R-50.}
% % \label{tab:ablations}
% \vspace{-0.8em}
% \end{table*}

 \Mingfei{In order to further validate the benefits of rectangular input, we evaluate the performance of X3D-M \cite{feichtenhofer2020x3d}, Uniformer-S \cite{li2022uniformer} and MViTv2-S \cite{li2022mvitv2} on non-square training resolutions and tested them on three portrait mode video benchmarks.
 % We also experiment with test crops in the same aspect ratios as the training crops and observe improvements. We evaluate X3D-M \cite{feichtenhofer2020x3d}, Uniformer-S \cite{li2022uniformer} and MViTv2-S \cite{li2022mvitv2} on their non-square training resolutions and report the results in Table~\ref{tab:ab_shape_attn_rect}.
 We find that the three models achieve higher accuracies on 3Massive and \abbrdataset with both crops in 256$\times$192 and 288$\times$192.
 On S100-PM, Uniformer-S and MViTv2-S achieve better testing results with $288\times192$ resolution, with FLOPs increased by around 15\% (47.5G vs 41.8 for Uniformer-S; 72.7G vs. 64.5G for MViTv2-S).
 % However, X3D-M and Uniformer-S show a slight decrease of 0.2\% in accuracies on S100-PM. 
 Note that FLOPs of 256$\times$192 are smaller than square 224$\times$224 (single clip inference cost: 40.6G vs. 41.8G for Uniformer-S; 62.9G vs. 64.5G for MViTv2-S).
 The performance boost further supports the potential benefits of rectangular input for video recognition in portrait mode.}

\section{The importance of temporal information}

% \Heng{We design several baselines to illustrate the importance of temporal information on portrait mode video. We will compare our dataset vs. popular benchmarks such as Kinetics-400 and Something-Something-V2.}

In this subsection, we investigate the importance of utilizing temporal information for portrait mode video recognition. We show that the \dataset is a valuable resource for evaluating video models in the challenging setting of portrait mode video recognition. 

We design two baselines with different temporal utilization approaches and extensively evaluate the models trained on Kinetics-400~\cite{kay2017kinetics}, 3Massiv~\cite{gupta20223massiv} and our \dataset. 
Specifically, we build our baselines with Uniformer-S and train the models with $224\times224$ crops. Uniformer-frames is constructed with image-based Uniformer-S and temporal aggregation of predicted logits using mean pool.
% like TSN~\cite{wang2016temporal}. 
It serves as a naive baseline since the temporal information is incorporated simply by merging the predicted logits across frames. For more advanced temporal correspondance, we train a video-based Uniformer-S endowed with self-attention on temporal dimension, building and learning temporal relations in different levels.

As shown in Table~\ref{tab:ab_temporal_info}, by leveraging temporal self-attention, Uniformer-S obtain accuracy improvement by 4.5\% and 4.6\% on Kinetics-400 and \dataset respectively. 
Interestingly, the 3Massiv dataset, most of which videos are in portrait mode, does not show as large of a performance gain from using temporal information as our \abbrdataset. 
% This may be because the 3Massiv dataset contains a high proportion of static videos, where the temporal information is less informative.
In contrast, our \dataset dataset shows a significant performance gain from using temporal information, attributable to its diverse collection of videos rich in intricate temporal dynamics.
% We argue that this is due to the diverse range of dynamic videos in PM-400, which includes many videos with complex temporal dynamics.
% attributable to its diverse collection of videos rich in intricate temporal dynamics.

% % See Table~\ref{tab:。/。//ab_temporal_info}.
% \begin{table}[t]
%     \small
%     \begin{center}
%         \input{tables/temporal_info}
%     \end{center}
%     % \vspace{-4mm}
%     \caption{Effect of utilizing temporal information for video recognition on different benchmarks.}
%     \label{tab:ab_temporal_info}
%     % % \vspace{-2mm}
% \end{table}

\section{The importance of the audio modality}

% \Heng{We design several baselines to illustrate the importance of audio modality on portrait mode video. We will compare our dataset vs. popular benchmarks such as Kinetics-400 and Something-Something-V2.}

In this section, we aim to explore the significance of audio information in portrait mode video recognition. To achieve this, we adopt the R3D-50~\cite{hara3dcnns} backbone trained on Kinetics-700~\cite{kinetics700} for spatio-temporal modeling and the VGG~\cite{harwath2018vision} model trained for sound classification~\cite{audioset} for audio modeling, following the practice in 3Massiv~\cite{gupta20223massiv}. We freeze the audio-visual backbones and train the classifier and multimodal fusion layers.

Our findings, as presented in Table~\ref{tab:ab_audio}, reveal that the model trained with audio consistently outperforms the model trained without audio on both the \abbrdataset and 3Massiv by approximately 2.4 points. This indicates that audio information plays a crucial role in portrait mode video recognition. Incorporating audio information can significantly enhance the performance of the model. We argue that audio cues can provide additional information about the subject's actions, emotions, and the surrounding environment, which poses unique challenges for video recognition in portrait mode.

% The results in this section suggest that audio information is an important modality for portrait mode video recognition. We encourage future researchers to explore the use of audio information for portrait mode video recognition tasks, such as action recognition, emotion recognition, and lip reading.

% See Table~\ref{tab:ab_audio}.
% \begin{table}[t]
%     \small
%     \begin{center}
%         \input{tables/audio_modality}
%     \end{center}
%     % \vspace{-4mm}
%     \caption{Comparison of different modalities with AVSlowFast.}
%     \label{tab:ab_audio}
%     % % \vspace{-2mm}
% \end{table}

% \begin{table}[t]
%     \small
%     \begin{center}
%         \input{tables/2stage_3massiv}
%     \end{center}
%     % \vspace{-4mm}
%     \caption{Comparison of different modalities with offline feature embeddings.}
%     \label{tab:ab_audio_3massiv}
%     % % \vspace{-2mm}
% \end{table}

% \section{The impact of video format to architecture design}

% \Heng{We can briefly discuss how does the change of video format from landscape mode to portrait mode change the design of video architectures, especially for vision transformers.}

\section{Discussions}
%\Heng{We can first reiterate our contributions here. Then comment on future directions that worth exploring for portrait mode video recognition.}

%\Heng{As we have describes in the previous sections, our paper is an invitation to researchers working on video understanding. We believe the research directions we discussed are critical to portrait mode video recognition, and our dataset can serve as a testbed to facilitate further works to better understand the new format of video data.}

In this work, we advocate conducting research on portrait mode videos. To this end, we introduce the \dataset dataset dedicated for portrait mode video recognition with a fine-grained taxonomy. We also make initial attempts to explore the specific properties of portrait mode videos, including their spatial bias, %es of portrait and landscape mode videos, 
and the optimal training and evaluation protocols, with effects of the temporal information and audio modality. % for portrait mode videos. 
We believe our dataset can serve as a testbed to facilitate further research such as novel architecture designs and multi-modality modeling on portrait mode videos.
%Our dataset can serve as a testbed to facilitate further works to better understand the new format of video data.
{
    \small
    \bibliographystyle{ieeenat_fullname}
    \bibliography{main}
}


\clearpage
\setcounter{page}{1}
\maketitlesupplementary

% We follow the official template from CVPR for organizing the supplementary material and the indexes of sections, figures and tables are continuous to the main submission for easy reference.

% \begin{table*}[!htbp]
% \centering
% \begin{tabular}{l cccccc}
% \hline

% \hline
% Dataset & Duration (h) & Avg.Dur. (s)  & Background & Classes & Videos \\ % & Origin
% \hline
% \hline
% \multicolumn{6}{c}{\textit{Hybrid orientation}}\\ % &  & & &  &
% \hline
% HMDB-51 & 12.22 & 6.5 &  Dynamic & 51 & 6,766 \\ % YouTube/Movies &
% UCF-101 & 26.67 & 7.2 &  Dynamic & 101 & 13,320 \\ % YouTube &
% ActivityNet-1.3 & 700.00 & 180  & Dynamic & 200 & 20,000 \\ % & YouTube
% Charades & 82.07 & 30 &  Dynamic & 157 & 9,848 \\ % 267 Homes &
% %AVA-v2.2 & 107.50 &  &  Dynamic & 80 & 386,000 \\ % 267 Homes &
% Kinetics-400 & 657.00 & 10  & Dynamic & 400 & 306,245 \\  % & YouTube
% \hline
% \multicolumn{6}{c}{ \textit{Portrai mode videos}} \\ % &  & &  & Representation
% \hline
% AutoTransition & 153.60 & 15.8 & Dynamic & 104 & 34,998 \\ % & YouTube
% Ours & 687.56 & 4  & Dynamic \& Static & 3094 & 618,800 \\ % & CapCut
% \hline

% \hline
% \end{tabular}
% \caption{Comparison with existing video datasets.}
% \label{tab:dataset2}
% \end{table*}

%%%%%%%%% BODY TEXT
\section{Implementation details}

% xxx

%-------------------------------------------------------------------------
\subsection{Training recipes}
% We present additional training details in addition to the frame resizing and input cropping strategy.
 In addition to the frame resizing and input cropping strategy, we provide additional training details. 
 Uniformer-S~\cite{li2022uniformer} is trained for 100 epochs with a learning rate of 0.0002, using a batch size of 96$\times$4 (the number of clips per GPU and the number of GPUs used for training). X3D-M~\cite{feichtenhofer2020x3d} is trained for 300 epochs with a learning rate of 0.2 and a batch size of 64$\times$4. MViTv2-S~\cite{fan2021mvit} is trained for 200 epochs with a learning rate of 0.0001 and a batch size of 32$\times$4.
 %We keep all the other settings, such as repeated augmentation~\cite{}, and learning rate decay identical as in their original papers. 
 We keep the other settings, such as repeated augmentation~\cite{hoffer2020augment}, and learning rate decay identical to those mentioned in their original papers.
 
\subsection{Testing recipes}
%In order to ensure fair comparison across all results presented in our paper, we conduct experiments using 224$\times$224 resolution, except for the cases specifically mentioned such as the rectangle crop input in Table 5 and spatial prior experiments in Section 4.2. 
To ensure a fair comparison of all results presented in our paper, we conducted experiments using a resolution of 224$\times$224, unless otherwise stated. For example, we used a rectangular crop input in Table 4 and conducted spatial prior experiments in Section 4.2.

%We employ identical test-time temporal and spatial augmentation as described in the original papers. Specifically, we use 10 temporal views with 3 spatial crops for X3D-M, 4 temporal views with 1 spatial crop for Uniformer-S, and 5 temporal views with 1 spatial crop for MViTv2-S. For more details on the augmentation methods used, we refer readers to the original papers.
We apply identical test-time temporal and spatial augmentation methods as described in the original papers. Specifically, for X3D-M, we use 10 temporal views with 3 spatial crops. For Uniformer-S, we use 4 temporal views with 1 spatial crop, and for MViTv2-S, we use 5 temporal views with 1 spatial crop. Readers who seek more detailed information on the augmentation techniques used can refer to the original papers.
%------------------------------------------------------------------------

\vspace{2mm}
\begin{figure}[t]
\begin{center}
   \includegraphics[width=1.0\linewidth]{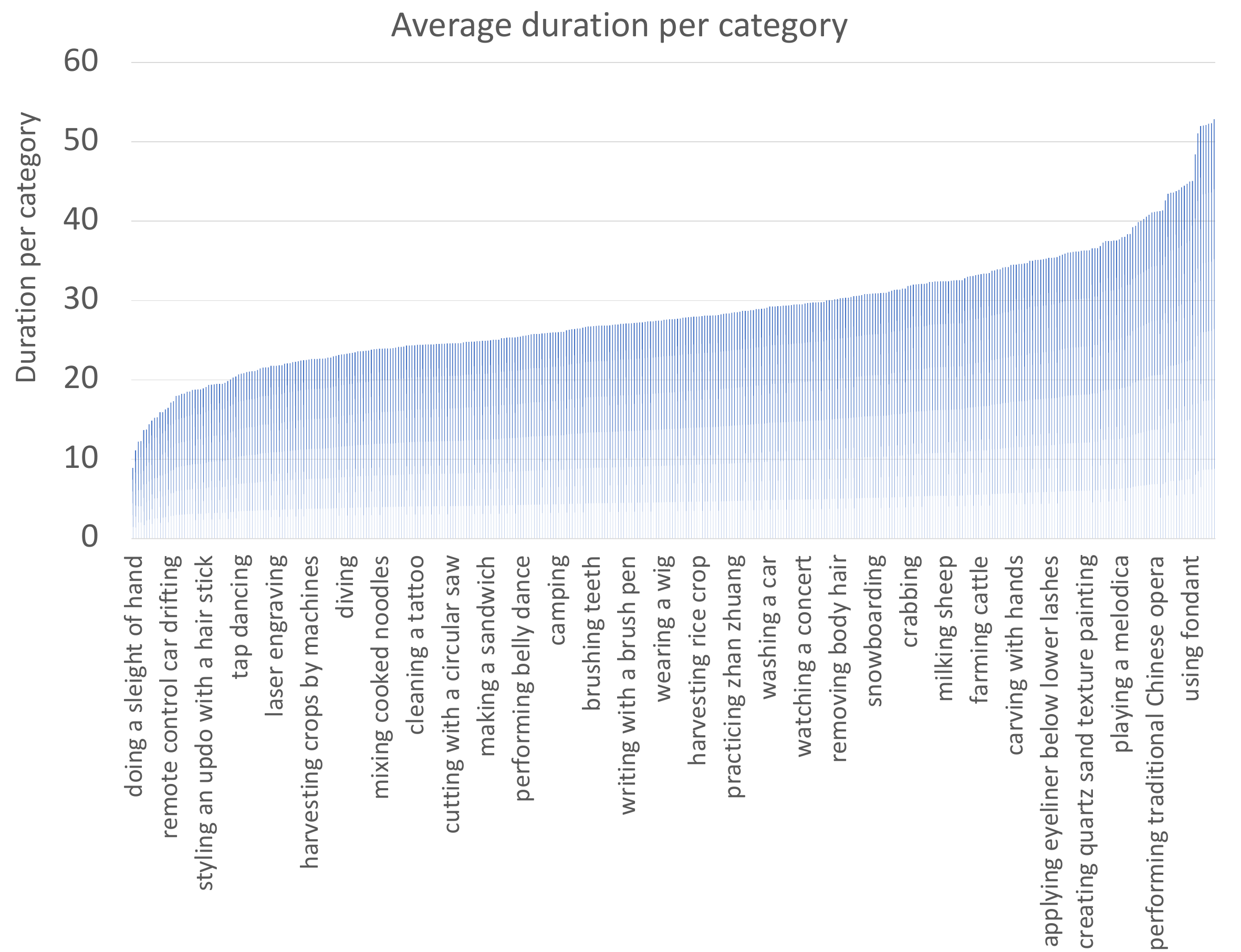}
\end{center}
\vspace{-1mm}
\caption{Average duration per category of \dataset. The horizontal axis indicates the category name, and the vertical axis represents the average duration per category.}
\label{fig:vis_duration}
\end{figure}

\section{\dataset}

\begin{figure}[t]
\begin{center}
   \includegraphics[width=1.0\linewidth]{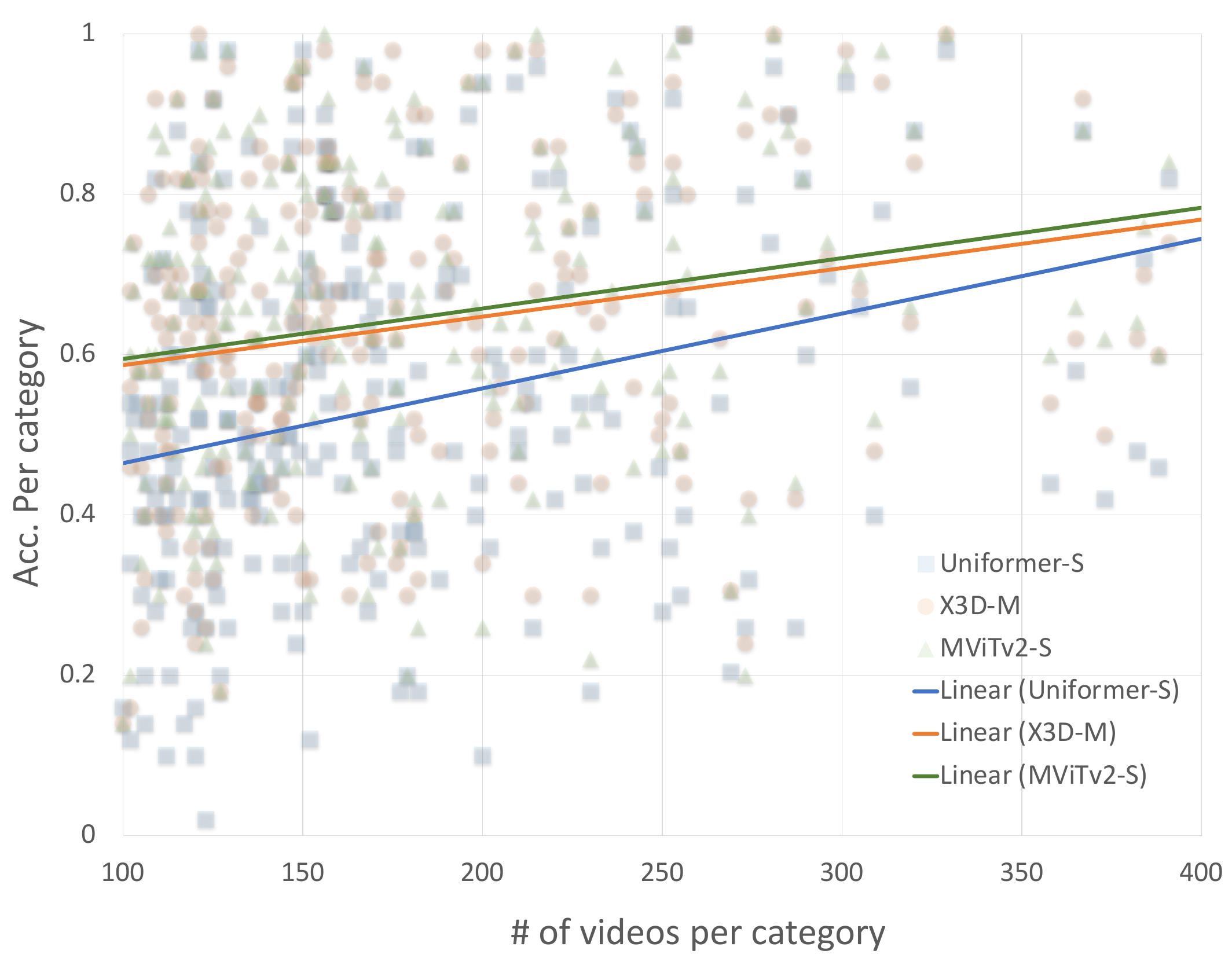}
\end{center}
\vspace{-2mm}
\caption{Accuracy per category of \dataset with Uniformer-S, X3D-M and MViTv2-S. The horizontal axis indicates the number of videos per category, and the vertical axis represents the accuracy of each respective category. The graph includes individual data points and a linear trend line.}
\vspace{-1mm}
\label{fig:vis_acc_cat}
\end{figure}

In addition to the dataset statistics presented in the main submission, we offer further insights into our dataset through visualizations. Figure~\ref{fig:vis_duration} displays the duration distribution of videos, while Figure~\ref{fig:vis_acc_cat} depicts accuracy per category.
These visualizations demonstrate that our dataset is well-balanced and suitable for training models that can handle a wide range of scenarios. Furthermore, we have created a webpage in \emph{demo} to highlight the diversity of our taxonomy and the unique characteristics of our videos. The \emph{demo} is included in the supplemental zip file.

%-------------------------------------------------------------------------
\subsection{Duration per category}

%The knowledge of the duration distribution of a video dataset can be crucial in developing models for video recognition tasks. 
%The duration distribution can be used to inform the selection of appropriate temporal scales for video analysis, such as the choice of window sizes or sampling rates. In the case of our dataset, Figure~\ref{fig:vis_duration} reveals a balanced distribution of video durations, providing important insights for optimizing video recognition models in portrait mode.
Understanding the duration distribution of a video dataset is critical in developing models for video recognition tasks. The distribution of durations can guide the selection of appropriate temporal scales for video analysis, including window sizes or sampling rates. For our dataset, Figure~\ref{fig:vis_duration} illustrates a balanced distribution of video durations, providing valuable insights for optimizing video recognition models in portrait mode.

%-------------------------------------------------------------------------
\subsection{Number of videos per category}

%Generally, having more training videos can lead to better performance in video recognition tasks due to increased prior knowledge. 
%Our experiments with three different architectures on our dataset further validate this observation.
%In Figure~\ref{fig:vis_acc_cat}, we plot the accuracy and number of videos for each category, along with a linear trend line. From the trend lines of the three models, it can be inferred that increased training video volume results in improved testing performance. 
In general, having a larger training video set can improve performance in video recognition tasks due to increased prior knowledge. Our experiments with three different architectures on our dataset further validate this observation.
Figure~\ref{fig:vis_acc_cat} displays the accuracy and number of videos for each category, along with a linear trend line. From the trend lines of the three models, we can infer that increased training video volume leads to improved testing performance.

%-------------------------------------------------------------------------
\subsection{Dataset glance}

%To offer a clearer view of our dataset, we have selected one video per category from 5 of our 9 domains. To view these videos, simply click on the \emph{index.html} webpage provided in the \emph{demo} directory. Clicking on the leaf node text will open a new page where the video can be played.
% To provide a more focused view of our dataset, we have hand-picked one video per category from five of our nine domains. These videos can be viewed by clicking on the \emph{index.html} webpage provided in the \emph{demo} directory. Clicking on the leaf node text will open a new page where the selected video can be played.

%Please note that the video samples have been downsampled to meet the file size limit of ICCV. However, we'll share links to the original videos for public release afterwards.
% Please note that the video samples provided in the \emph{demo} directory have been downsampled to meet the file size limit of CVPR. However, we will share links to the original videos for public release afterwards.
To concisely present our dataset, we have selected one representative video from each category across four of our nine domains. Access to these videos is available through the \emph{index.html} page in the \emph{demo} directory, where clicking on the leaf node text redirects to a webpage for video playback.

It's important to note that for CVPR's reviewing process, we provide 40 downsampled video files in the \emph{demo} directory, adhering to the conference's supplementary file size constraints. However, our dataset does not store or distribute raw videos. Instead, for public release, we will provide links to the video sources, requiring users to download the videos themselves. This approach ensures compliance with distribution guidelines while facilitating ease of access and review.

\section{Zero-shot video recognition}

% \Mingfei{We propose additional 100 categories for zero-shot video recognition.}
% In this section, we introduce a zero-shot portrait-mode video recognition dataset. Beyond our initial 400 categories, we have curated an extra 100 categories specifically for zero-shot video recognition. These categories encompass a broad spectrum of aspects, including vehicles, daily life, and art.

% We have previously discussed the unique characteristics of spatial prior, the importance of temporal information, and the value of the audio modality in portrait mode video recognition. These features are inherited by our zero-shot dataset, which we believe to be a valuable resource for the research community. This dataset not only extends the original zero-shot video recognition settings but also introduces additional challenges, making it an intriguing subject for future research.

% To provide a starting point for research on this dataset, we establish our baselines using the CLIP model with a mean pooling temporal aggregation approach, as implemented in CLIP4clip. We also present the performance of XCLIP for comparison. We hope that our zero-shot portrait mode video dataset will inspire and facilitate further exploration in this field.

In this section, we expand our \dataset with a zero-shot portrait-mode video recognition dataset, introducing 100 newly curated categories specifically designed for zero-shot recognition. Our dataset embodies critical aspects of portrait mode video recognition, notably spatial priors, temporal information, and audio modality.
% , pivotal in portrait-mode analysis. 
This enrichment not only challenges but also broadens the horizons of zero-shot video recognition research.

\begin{table}[h]
    \small
    \begin{center}
        
\setlength{\tabcolsep}{2.3mm}
\footnotesize
\begin{tabular}
{c c  c c }
\toprule

Model & Pretrain & Accuracy & Views\\
% \arrayrulecolor{white}\cline{1-4}
% \arrayrulecolor{black}\cline{1-4}
% \arrayrulecolor{black}\cline{1-4}
% \arrayrulecolor{white}\cline{1-4}
% \arrayrulecolor{white}\hline
% \arrayrulecolor{black}\hline
% \arrayrulecolor{white}\hline
\midrule
CLIP~\cite{radford2021learning} & CLIP400M & 50.1 & $1\times 1$ \\
% CLIP  & CLIP400M  & 50.5 & $1\times 4$ \\
% CLIP  & CLIP400M  & 48.7 & $3\times 1$ \\
% CLIP  & CLIP400M  & 48.8 & $3\times 4$ \\
\midrule
% \arrayrulecolor{white}\cline{2-4}
% \arrayrulecolor{white}\cline{2-4}
% \arrayrulecolor{black}\cline{2-4}
% \arrayrulecolor{white}\cline{2-4}
% \arrayrulecolor{white}\cline{2-4}
X-CLIP~\cite{XCLIP} & CLIP400M+K600 & 52.5 & $1\times 1$\\
% X-CLIP~\cite{XCLIP} & CLIP400M+K600 & 53.7 & $1\times 4$\\
% X-CLIP~\cite{XCLIP} & CLIP400M+K600 & 53.5 & $3\times 1$\\
X-CLIP~\cite{XCLIP} & CLIP400M+K600 & 54.6 & $3\times 4$\\
% \arrayrulecolor{white}\cline{2-4}
% \arrayrulecolor{white}\cline{2-4}
% \arrayrulecolor{black}\cline{2-4}
% \arrayrulecolor{white}\cline{2-4}
% \arrayrulecolor{white}\cline{2-4}
% UMT~\cite{li2023unmasked} & x & x & x & x \\
% \arrayrulecolor{white}\hline
% \arrayrulecolor{black}\hline
% \arrayrulecolor{white}\hline
\bottomrule

\end{tabular}
    \end{center}
    \vspace{-4mm}
    \caption{Performance on our reserved zero-shot subset. Views during inference are shown by
the multiplication of \# of spatial crops and \# of temporal views.}
    \label{tab:zero-shot}
    \vspace{-2mm}
\end{table}

As shown in Table~\ref{tab:zero-shot}, to facilitate initial explorations, we establish baselines using the CLIP model~\cite{radford2021learning} with a mean pooling temporal aggregation method. Additionally, we present a comparative analysis with XCLIP~\cite{XCLIP}'s performance. We aim for this dataset to drive forward research and innovation in this domain.

\section{Broader impact}

\noindent \textbf{Data Limitations and Ethical Considerations.} We do not store or distribute videos; users must obtain them from the original sources. Furthermore, our careful manual annotation process is designed to prevent any ethical or legal issues.

\noindent \textbf{Human Rights in Annotation Process.} We have carefully organized the annotation task to guarantee reasonable workloads and just remuneration for annotators, adhering to human rights principles.

\noindent \textbf{Scope of Conclusions.} It is essential to be aware that experiments and data, including ours, may only be a small part of the whole picture. Nevertheless, due to the extensive range of data, such as 3Massiv, S100 and \dataset, that we have used in our experiments, we are confident that our results offer a reliable comprehension that can be applied to portrait-mode video analysis. Although these discoveries are particular to the data we have examined, they offer considerable insight into the wider area of video analysis.

\noindent \textbf{Future Research and Development.}
Aligned with our commitment to the research community and in adherence to CVPR guidelines, we will release both our code and dataset. This is intended to encourage further research and enable others to build upon our work. 
% Although our current experiments require up to 8$\times$2 A100-GPUs, we are aware this may be a limitation for some institutions. 
% Consequently, we plan to focus future efforts on adapting these experiments to be compatible with a single node of 8 A100 GPUs. 
% It's important to note that fitting the experiments within an 8 GPU framework is not the primary focus of this paper, but we consider it a crucial step towards making our research more accessible and inclusive for a wider array of research groups.

\end{document}